\documentclass[preprint,12pt]{elsarticle}
\usepackage[top=2cm, bottom=2cm, left=3cm, right=3cm]{geometry}
\usepackage{amsmath}
\usepackage{amssymb}
\usepackage{graphicx}
\usepackage{hyperref}
\usepackage{comment}
\usepackage{pdfpages}
\usepackage{booktabs}
\usepackage{array}
\usepackage{xcolor}
\usepackage{caption}
\usepackage{multirow}
\usepackage{siunitx}
\usepackage{parskip}
\usepackage{algorithm}
\usepackage{algorithmic}
\usepackage{setspace}
\usepackage{pdflscape}
\usepackage{tabularx} 

\newsavebox\CBox
\def\textBF#1{\sbox\CBox{#1}\resizebox{\wd\CBox}{\ht\CBox}{\textbf{#1}}}

\begin{document}

\begin{frontmatter}
\title{Information-Theoretic Dual Memory System for Continual Learning}

\author[inst1]{RunQing Wu}
\author[inst2]{KaiHui Huang}
\author[inst3]{HanYi Zhang}
\author[inst5]{QiHe Liu}
\author[inst6]{GuoJin Yu}
\author[inst7]{JingSong Deng}
\author[inst4]{Fei Ye\texorpdfstring{\corref{cor1}}{}}
\ead{feiye@uestc.edu.cn}

\cortext[cor1]{Corresponding author}

\address[inst1]{School of Mechanical Engineering, Huazhong University of Science and Technology, China}
\address[inst2]{School of Software Engineering, University of Science and Technology of China}
\address[inst3]{School of Computation, Information and Technology, Technische Universität München, Germany}
\address[inst5]{School of Information and Software Engineering, University of Electronic Science and Technology of China}
\address[inst6]{School of Information and Software Engineering, University of Electronic Science and Technology of China}
\address[inst7]{CAS Key Laboratory of Mental Health, Institute of Psychology, Chinese Academy of Sciences, China}
\address[inst4]{School of Information and Software Engineering, University of Electronic Science and Technology of China}

\date{\today}

\begin{abstract}
Continuously acquiring new knowledge from a dynamic environment is a fundamental capability for animals, facilitating their survival and ability to address various challenges. This capability is referred to as continual learning, which focuses on the ability to learn a sequence of tasks without the detriment of previous knowledge. A prevalent strategy to tackle continual learning involves selecting and storing numerous essential data samples from prior tasks within a fixed-size memory buffer. However, the majority of current memory-based techniques typically utilize a single memory buffer, which poses challenges in concurrently managing newly acquired and previously learned samples. Drawing inspiration from the Complementary Learning Systems (CLS) theory, which defines rapid and gradual learning mechanisms for processing information, we propose an innovative dual memory system called the Information-Theoretic Dual Memory System (ITDMS). This system comprises a fast memory buffer designed to retain temporary and novel samples, alongside a slow memory buffer dedicated to preserving critical and informative samples. The fast memory buffer is optimized employing an efficient reservoir sampling process. Furthermore, we introduce a novel information-theoretic memory optimization strategy that selectively identifies and retains diverse and informative data samples for the slow memory buffer. Additionally, we propose a novel balanced sample selection procedure that automatically identifies and eliminates redundant memorized samples, thus freeing up memory capacity for new data acquisitions, which can deal with a growing array of tasks. Our methodology is rigorously assessed through a series of continual learning experiments, with empirical results underscoring the effectiveness of the proposed system.
\end{abstract}

\end{frontmatter}

\section{Introduction}
\noindent

In practical scenarios and applications, data samples are typically provided in a sequential manner, precluding the possibility of accessing all data simultaneously. This form of learning framework within computer science is identified as continual learning \cite{LifeLong_review}. While contemporary deep learning models have demonstrated remarkable effectiveness in a single, static dataset, their application within a continual learning framework poses significant challenges due to the potential loss of previously acquired knowledge when learning new tasks. Such a phenomenon of performance degeneration is referred to as catastrophic forgetting \cite{LifeLong_review}.

Recent studies in the field of continual learning have introduced five distinct methodologies to mitigate catastrophic forgetting, namely: regularization-based strategies \cite{EWC}, memory or experience replay techniques \cite{RainbowMemory}, optimization-based methods, representation-based methods and architecture-based strategies \cite{ProgressiveNN}. Among these methodologies, the memory replay technique stands out as both straightforward and widely adopted in continual learning. It operates via a compact memory buffer that retains a limited number of past data samples \cite{RainbowMemory}. When engaging in new task learning, these retained samples are integrated with new data for model training, thereby the model's performance heavily relies on the quality of the memorized samples. The regularisation-based approach, conversely, seeks to regulate the optimization process of the model to attain optimal performance across both prior and current tasks \cite{CLNullSpace}. This technique incorporates an additional regularization term into the main objective function, aiming to penalize significant alterations in many critical network parameters. Regularisation strategies are well-equipped to tackle traditional continual learning challenges and are also applicable to online continual learning scenarios \cite{OnlineLearning} by combining them with memory-based approaches. Another category of continual learning methods is denoted as dynamic expansion models \cite{LifelongInfinite}, which inherently enhance model capacity by adding new hidden layers and nodes within the existing network architecture. In contrast to static network models \cite{ContinualPrototype}, dynamic expansion models present several advantages, including scalability and robust generalization performance for previously learned tasks. Nonetheless, a significant drawback of this approach is the increased computational complexity and storage requirements, posing challenges in real-world contexts where devices and machines have severe resource constraints.

The Complementary Learning Systems (CLS) theory elucidates an important biological mechanism within the brain, in which the information is processed through two interdependent systems: a gradual learning mechanism designed to retain novel skills and experiences from a dynamically evolving environment, and a rapid learning mechanism aimed at efficiently distilling essential information from prior experiences to reinforce acquired knowledge \cite{HumanLerningSystem}. Inspired by the results from the CLS framework, many studies have developed several innovative continual learning systems \cite{DualCL,CLFast}. Nonetheless, these approaches either necessitate the incorporation of an auxiliary neural network \cite{DualCL} or the implementation of an additional optimization procedure to streamline the training process \cite{CLFast}, thereby leading to increased computational overhead and memory requirements.

Motivated by the results derived from the CLS theory framework and considering the weaknesses in existing methodologies, we propose an innovative continual learning approach that implements both fast and slow learning mechanisms from a memory-centric viewpoint. Specifically, we propose to split the whole memory system into two parts~: a fast memory buffer designed to randomly capture recent data samples along with a slow memory buffer intended to retain essential and more informative data derived from past experiences. This memory design implements the rapid and gradual learning mechanisms within the CLS theory framework by facilitating the model's ability to assimilate both novel and previously critical data samples without incurring additional training expenses. Furthermore, to selectively store high-calibre data samples within the slow memory buffer, we introduce a novel Information-Theoretic Memory Optimization (ITMO) strategy that leverages information theory techniques to identify more informative data samples. This ITMO approach designs a novel objective for sample selection that balances two factors: diversity and similarity. The diversity term evaluates the representativeness of the selected subset, where higher diversity signifies that the samples contain a broad spectrum of semantic information. Conversely, the similarity term assesses the closeness between the selected subset and the original training set, with lower discrepancies indicating that the selected samples represent more statistical relevance to the original training dataset. The proposed information cost function implements these two attributes using the second-order Renyi entropy and Cauchy-Schwarz (CS) divergence loss terms, enabling to select and store more informative data samples in the slow memory buffer.

In addition, we propose to improve the model's generalization performance by maximizing the use of the memory space by retaining as many samples as feasible, thereby providing sufficient training samples for the model training. Nonetheless, upon the completion of a task learning phase, the proposed memory system lacks the capacity to accommodate the data samples from the current task learning, leading to catastrophic forgetting in the subsequent task learning. To mitigate this issue, we introduce an innovative sample selection approach that systematically eliminates numerous redundant memorized samples after each task switch, thereby freeing up memory space for preserving novel data samples for the new task learning. Furthermore, this proposed sample selection strategy guarantees an equitable distribution of memorized samples across each category following the sample removal procedure, effectively tackling the imbalanced continual learning problem. Our dual memory framework is designed for seamless integration, allowing it to be effortlessly implemented within existing continual learning models to enhance their efficacy. We conduct a comprehensive series of experiments to assess the effectiveness of our proposed methodology, with empirical results indicating that it achieves cutting-edge performance across diverse continual learning scenarios.

We summarize our main contributions in the following~:
\begin{enumerate}
\item \textbf{The Dual Memory System~:} Inspired by the results from the CLS theory framework, we propose a novel dual memory system to preserve both new and previous critical data samples. The proposed memory system is plug-and-play and can be applied to various continual learning settings without significant modifications.
    \item \textbf{Information-Theoretic Sample Selection~:} We propose a novel memory optimization approach from the information theory perspective. The proposed approach introduces an information cost function to evaluate the quality of each chosen sample, which helps selectively store many representative and informative data samples in the slow memory buffer.
    \item \textbf{The Balanced Memory Optimization~:} We propose a novel sample selection approach to automatically select and remove many redundant memorized samples, which can provide enough memory capacity to store new data samples. In addition, the proposed sample selection approach can ensure the 
sample balance and diversity of the slow memory buffer after the sample removal process.
    \item \textbf{Experiments~:} We construct a series of experiments on various continual learning settings and compare our approach with a broad range of continual learning methods. The empirical results show that using the proposed dual memory system in the existing continual learning models can improve their performance.
\end{enumerate}

\section{Background and Related Work}
\label{relatedWork}

Catastrophic forgetting is a significant challenge in continual learning (CL), where models would lose all previously learnt knowledge when training on a new task \cite{LifeLong_review}. To address this, various continual learning technologies have been proposed, including regularization-based methods \cite{FlattingCL,BooVAE,ReusableCL,Afec}, memory replay-based methods\cite{GradientEpisodic,AGEM,DER,rebuffi2017icarl,prabhu2020gdumb,riemer2019learning}, optimization-based methods \cite{CLBlurry,NoSelection,CLMutual,jha2024npcl,LossDecouplingCL,Closer_ContinualLearning,GCR,PromptingCL}, representation-based methods\cite{madaan2021representational,DualCL,mehta2023empirical,cong2020gan}, and architecture-based methods \cite{ForgetFree,LearnAdd,ProgressiveNN,Error_driven,TFCLTransformer,LifelongInfinite}. In this section, we firstly review these approaches focusing on their mathematical foundations and contributions to mitigating forgetting. We will then briefly introduce current approaches that leverage information-theoretic methods in CL. “We summarize the representative methods listed above in Table.~\ref{tab:cl_methods}.”

\sloppy
\begin{table}[t]
\centering
\caption{Summary of Continual Learning Methods.}
\fontsize{9}{11}\selectfont
\centering
\renewcommand{\arraystretch}{1.5}
\setlength{\tabcolsep}{2mm} 
\resizebox{\textwidth}{!}{ 
\begin{tabular}{@{}l p{7cm} p{4.2cm}@{}}
\toprule
\textbf{Method Category} & \textbf{Mathematical Principles} & \textbf{Representative} \\
\midrule
Regularization-Based 
& \raggedright $\mathcal{L}_{\text{total}} = \mathcal{L}_{\text{new}} + \lambda \sum_i F_i (\theta_i - \theta_i^*)^2$, where $\lambda$ controls regularization strength; $F_i$ is derived from the Fisher Information Matrix; penalizes changes to important parameters $\theta_i$. 
& EWC \cite{EWC}, SI \cite{Continual_Learning}, LwF \cite{Lwf}, KD \cite{KD_Review} \\
Replay-Based 
& \raggedright $\mathcal{L}_{\text{total}} = \mathcal{L}_{\text{new}} + \alpha \mathcal{L}_{\text{replay}}$, where $\alpha$ balances the loss of current task and replayed examples; $\mathbf{x}_{\text{gen}} = G(\mathbf{z}; \theta_G)$ generates replay data.
& ER \cite{wang2022hierarchical}, GR \cite{shin2017continual}, RS \cite{vitter1985random}, DGR \cite{shin2017continual}, MeRGAN \cite{MemoryReplayGAN} \\
Optimization-Based 
& \raggedright Minimize $\mathcal{L}$ while ensuring $\nabla L_{\text{new}}(\theta) \cdot \nabla L_{\text{old}}(\theta) \geq 0$; aligns gradients or projects onto orthogonal subspaces to avoid interference. 
& GEM \cite{GradientEpisodic}, OGD \cite{farajtabar2020orthogonal}, GPM \cite{saha2021gradient}, MAML \cite{finn2017model}, OML \cite{javed2019meta}, La-MAML \cite{gupta2020look} \\
Representation-Based 
& \raggedright Learn shared features $h(\mathbf{x})$ that minimize task interference; objective $\min \mathcal{L}_{\text{SSL}} + \mathcal{L}_{\text{task}}$ combines self-supervised and supervised learning losses.
& SSL \cite{pham2021lump}, DualNet \cite{DualCL}, EWC \cite{EWC}, IncCLIP \cite{incclip2022} \\
Architecture-Based 
& \raggedright Dynamically expand parameters $\Theta = \bigcup_{t=1}^T \Theta_t$, adding task-specific components while preserving previously learned knowledge.
& PNNs \cite{PNN}, DEN \cite{Lifelong_expandable}, PathNet \cite{fernando2017pathnet}, NP \cite{han2015deep}, PS \cite{serra2018overcoming} \\
Information-Theoretic 
& \raggedright Maximize entropy $H(\mathbf{X})$ to ensure diversity, minimize mutual information $I(\mathbf{X}; \mathbf{Y})$ to reduce redundancy and maintain relevant features. 
& SUR \cite{sun2022information}, OL \cite{song2023infocl}, BM \cite{hierarchically2023structured}, CL \cite{fcil2023fewshot} \\
\bottomrule
\end{tabular}
}\label{tab:cl_methods}
\end{table}

\subsection{Regularization-Based Methods}
\noindent Regularization-based methods mitigate catastrophic forgetting by introducing explicit regularization terms into the loss function, balancing the learning between old and new tasks\cite{mao2021continual,Lwf,Continual_Learning}. These methods often require storing a reference copy of the previous model to guide the regularization process. Regularization-based approaches can be broadly categorized into weight regularization and function regularization.

Weight regularization focuses on constraining parameter updates to preserve knowledge from previous tasks. For instance, Elastic Weight Consolidation (EWC) \cite{EWC} introduces a regularization term $ \sum_i \frac{\lambda}{2} F_i (\theta_i - \theta_i^*)^2 $, where $ \lambda $ is a hyperparameter controlling the regularization effects, $F_i$ is derived from the Fisher Information Matrix, and $ \theta_i^*$ represents the optimal parameters from previous tasks. This approach minimizes significant changes in critical parameters during the learning of new tasks, effectively stabilizing key parameters. Synaptic Intelligence (SI) \cite{Continual_Learning} is another method that estimates parameter importance by tracking their contributions to the total loss over time, adjusting updates to preserve crucial knowledge. Function regularization, on the other hand, aims to maintain consistent model outputs by ensuring that predictions for new tasks align with those of the old model. This is often achieved through knowledge distillation (KD) \cite{KD_Review}, where the previous model serves as a teacher. The Learning without Forgetting (LwF) \cite{Lwf} is a popular teacher-student framework in which the distance between the teacher's and student's outputs for a given new data is minimized during the training process, which can relieve forgetting.

In conclusion, regularization techniques provide robust approaches to stabilize essential parameters or ensure consistent outputs, rendering them particularly advantageous in situations where it is vital to uphold existing knowledge. Nonetheless, these methods might necessitate extra computational resources to sustain and refresh the reference model.

\subsection{Replay-Based Methods}
\noindent Besides regularization techniques, replay-based methods also play a crucial role in mitigating catastrophic forgetting by incorporating data from previous tasks during the learning of new ones\cite{wang2022hierarchical,ji2021coordinating}. This is typically achieved by combining the loss functions of old and new tasks to form a total loss that balances the importance of both. Replay-based methods are primarily divided into experience replay and generative replay.

Experience replay involves maintaining a buffer of selected samples from previous tasks, which are replayed during the training of new tasks. The challenge lies in optimizing the selection and storage of these samples due to limited memory capacity. Techniques like Reservoir Sampling\cite{vitter1985random} provide basic solutions by selecting representative samples, while more advanced methods dynamically adjust the buffer content to capture the most critical data from previous tasks.

Generative replay, or pseudo-rehearsal, addresses the storage limitation by using a generative model \( G(\mathbf{z}; \theta_G) \) to create synthetic data that mimics past tasks, where $\theta_G$ denotes the parameters of the generator. During training, this model generates samples \( \mathbf{x}_\text{gen} = G(\mathbf{z}; \theta_G) \) that are used in place of real data from earlier tasks. Frameworks like Deep Generative Replay (DGR)\cite{shin2017continual} integrate the training of both the task model and the generative model, preserving previous knowledge. Methods such as MeRGAN\cite{MemoryReplayGAN} further improve this by ensuring consistency between the synthetic data produced over time, reducing potential drift in the generated distributions.

In summary, replay-based methods are particularly effective in environments with high task variability. However, these methods may require significant storage for experience replay or additional computational power for generative models, especially when dealing with complex datasets.

\subsection{Optimization-Based Methods}
Optimization-driven techniques alleviate the issue of forgetting by meticulously modifying the optimization framework during training to safeguard knowledge from antecedent tasks. These techniques frequently entail adjustments to gradient updates to avert interference from recently introduced tasks. A prevalent strategy is gradient projection, which aligns the present gradient updates with those of preceding tasks or projects them onto orthogonal subspaces to minimize conflict. Gradient Episodic Memory (GEM)\cite{GradientEpisodic} is a classic optimization approach, which ensures that the gradient associated with a new task $ \nabla L_{\text{new}}(\theta) $ remains congruent with earlier gradients $ \nabla L_{\text{old}}(\theta) $, thereby upholding the criterion $ \nabla L_{\text{new}}(\theta) \cdot \nabla L_{\text{old}}(\theta) \geq 0 $. This approach aids in averting the deterioration of previously acquired knowledge. Additional methodologies, such as Orthogonal Gradient Descent (OGD)\cite{farajtabar2020orthogonal} and Gradient Projection Memory (GPM)\cite{saha2021gradient}, enhance this by projecting the gradient updates onto subspaces that are orthogonal to the significant gradient vectors of past tasks, thus preserving crucial information while facilitating new learning tasks.

Meta-learning, often referred to as "learning to learn" represents a pivotal concept in the machine learning field. The main goal of using the meta-learning is to improve the model's responsiveness to novel tasks by leveraging previous experiences. Model-Agnostic Meta-Learning (MAML)\cite{finn2017model} focuses on optimizing parameters that enable swift fine-tuning for emerging tasks while minimizing the risk of knowledge retention loss. Methodologies such as OML\cite{javed2019meta} and La-MAML\cite{gupta2020look} harmoniously merge meta-learning with continual learning, blending gradient modifications with experience replay to strike a balance between the preservation of established knowledge and the assimilation of new information.

In conclusion, optimization-based techniques provide a resilient framework for continual learning by regulating gradient adjustments to maintain previously acquired knowledge or by utilizing meta-learning to improve flexibility. Nevertheless, these approaches may demand significant computational resources, which makes them particularly applicable in small-size machines and devices.

\subsection{Representation-Based Methods}
Representation-based approaches mitigate the issue of forgetting by cultivating common representations that are transferable across various tasks, which in turn minimizes interference and safeguards knowledge retention. Generally, the lower layers of the network are responsible for acquiring these shared features $ h(\mathbf{x})$, whereas the specialized classifiers $ f_t(h(\mathbf{x}))$ situated in the upper layers are tasked with executing predictions for each individual task $t$.

Recent innovations have integrated self-supervised learning (SSL) and extensive pre-training to significantly improve representation learning. SSL methodologies, such as those utilized in LUMP\cite{pham2021lump} and Co2L\cite{Co2l}, leverage contrastive loss to establish robust representations that are resilient against forgetting. Furthermore, dual-network frameworks such as DualNet \cite{DualCL} and CL-SLAM\cite{parisi2022clslam} synergize supervised and self-supervised learning to achieve a balance between generalization and stability. The efficacy of large-scale pre-training is also notable, as models that are trained on comprehensive datasets tend to exhibit greater resilience to forgetting and demonstrate superior knowledge transfer capabilities to novel tasks. Nonetheless, a prominent challenge persists in the adaptation of these pre-trained representations to new tasks while preserving their wide applicability. This issue is navigated through various strategies that depend on whether the pre-trained features remain static or are dynamically adjusted during new task learning. Additionally, continual pre-training and meta-training methodologies allow models to incrementally refine their representations as fresh data is introduced. For example, merging techniques like Barlow Twins with Elastic Weight Consolidation (EWC)\cite{zbontar2021barlow} facilitates learning from incremental data, thereby enhancing the model's aptitude for accommodating new tasks, whereas methodologies such as IncCLIP\cite{incclip2022} perpetually update multi-modal models through the replay of generated samples.

In conclusion, representation-based approaches offer a strong framework for continual learning through the creation of shared representations that effectively generalize across various tasks. Although they are efficient, these methods necessitate meticulous oversight of the equilibrium between stability and adaptability, particularly when addressing a wide range of diverse tasks.

\subsection{Architecture-Based Methods}
Architecture-driven techniques tackle the forgetting problem in continual learning by adaptively altering the model's structure when learning new tasks. This methodology often consists of augmenting the network with additional layers or modules tailored to each task, thereby safeguarding the newly acquired knowledge. Mathematically, this is realized by dynamically building the task-specific parameters $\theta_t$ for learning a new task. The overall model's parameters can be formulated as the union of task-specific parameter sets $ \Theta = \bigcup_{t=1}^T \Theta_t $, in which each $ \Theta_t $ preserves the information for a specific task.
A most popular architecture-based approach is Progressive Neural Networks (PNNs)\cite{PNN}, where new network channels are added for each task learning. Such a mechanism prevents catastrophic forgetting and also enables learning a growing number of tasks. However, the PNNS can lead to considerable computational complexity and memory costs when learning a long sequence of tasks. Dynamically Expandable Networks (DEN)\cite{Lifelong_expandable} and PathNet\cite{fernando2017pathnet} offer more refined solutions by selectively growing the network only when necessary, based on task complexity, and reusing existing components to handle different tasks. Techniques like network pruning\cite{han2015deep} and parameter sharing\cite{serra2018overcoming} have been explored to mitigate these issues by reducing redundancy and improving efficiency.

In conclusion, architecture-based strategies present a robust framework for continual learning by effectively creating new parameters to adapt to a new task. Nonetheless, the model's size and computational complexity are the primary drawback of the architecture-based methods.

\subsection{Information-Theoretic Approaches in Continual Learning}
Recent studies have integrated information-theoretic principles to improve memory rehearsal strategies in continual learning\cite{sun2022information,song2023infocl,hierarchically2023structured,fcil2023fewshot,itframework2023causal}. By employing criteria such as "surprise" and "learnability," they aim to retain diverse and relevant information while balancing computational efficiency\cite{sun2022information,song2023infocl}. These methods leverage online selection strategies, Bayesian models, and contrastive learning to maintain a representative subset of data, which helps mitigate catastrophic forgetting and enhances model robustness in imbalanced or dynamic learning environments\cite{hierarchically2023structured,fcil2023fewshot}. Entropy, a measure that assesses the uncertainty or randomness of a dataset, is crucial for ensuring diversity during sample selection. Moreover, mutual information evaluates the shared information between variables, which is vital for identifying and maintaining relevant features while minimizing redundancy. Although information-theoretic methods have been extensively utilized in domains such as feature selection, clustering, and anomaly detection, their potential in continual learning still have rooms to improve. By leveraging these principles, the effectiveness of memory buffers in rehearsal-based strategies can be significantly enhanced, leading to a comprehensive and representative data subset for continual learning applications.

To summarize, although each of these methodologies presents distinct benefits in tackling the forgetting problem in continual learning, they are accompanied by trade-offs relating to computational resources, memory demands, and model intricacy. The approach we propose is grounded in these principles, incorporating information-theoretic concepts into both data subset selection and memory buffer optimization, with the goal of improving the model’s capacity for addressing forgetting in continual learning.

By conducting thorough experimental assessments, we corroborate the effectiveness of our methodology in enhancing the performance of continual learning models, especially in scenarios characterized by imbalanced data. Our findings introduce a comprehensive framework for refining memory buffer strategies and strengthening the resilience of CL models against the phenomenon of catastrophic forgetting.

\begin{figure}[t]
    \centering
    \includegraphics[trim=0.25cm 4.2cm 0.5cm 3.3cm, clip, width=1.0\textwidth, page=1]{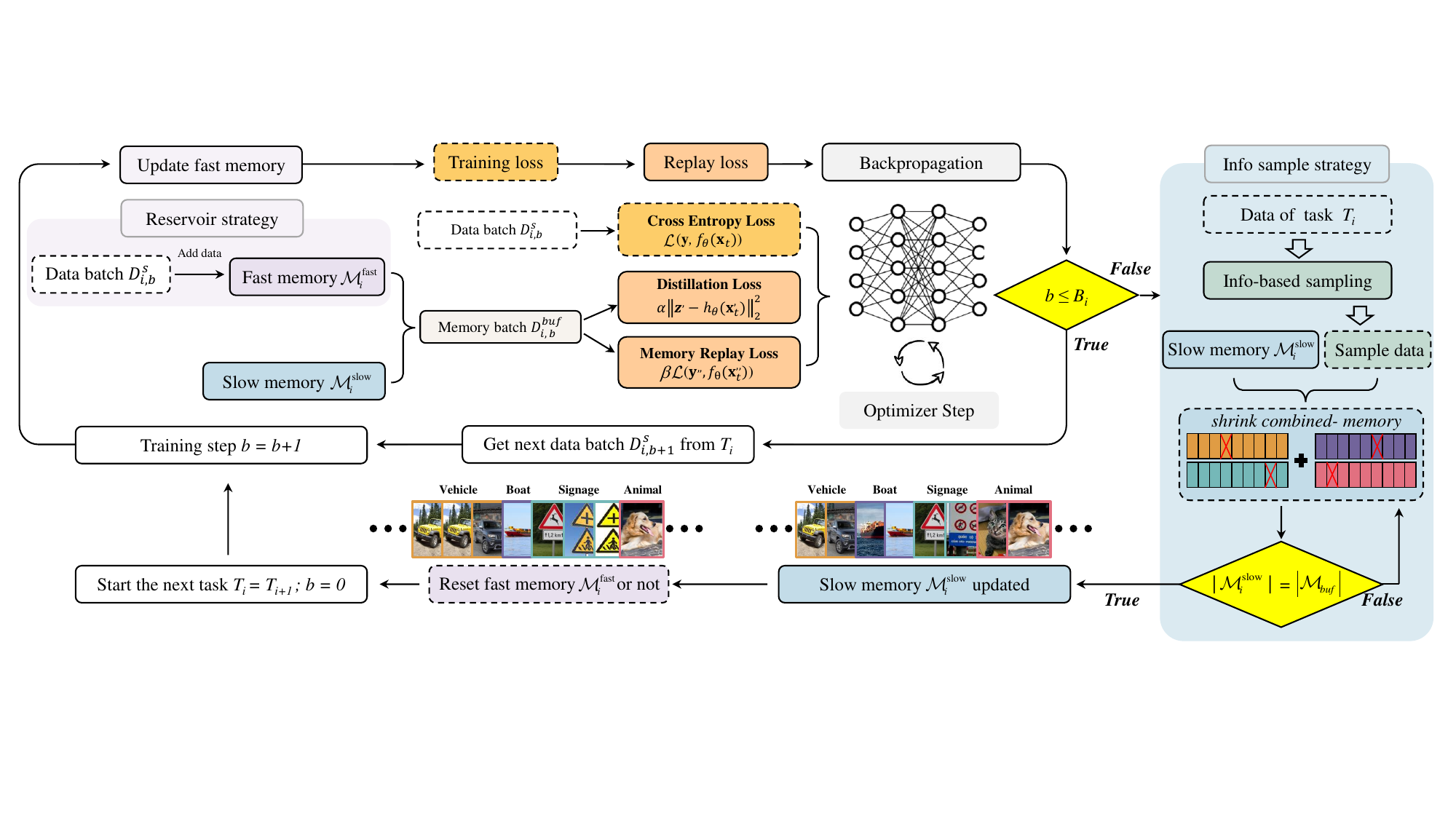}
    \caption{The training procedure for the proposed dual memory system consisting of a fast and slow memory buffer, respectively. The fast memory buffer continually stores new data samples or replaces old memorized samples with new ones via reservoir sampling. Assuming that the model was trained at the $i$-th task learning ($i>1$), we first remove samples from the slow memory buffer using Eq.~\eqref{sampleRemove}. Then we optimize the sample selection probability using Eq.~\eqref{weightOptimization2} and add the data samples from $D^s_i$ according to the selection weight vector ${\bf w}$. 
    }
\label{fig:training}
\end{figure}

\section{Methodology}
\label{method}

\begin{table}[t]
\caption{Description of Important Mathematical Notations}
    \centering
\fontsize{9}{11}\selectfont
\renewcommand{\arraystretch}{1.1} 
\setlength{\tabcolsep}{4mm}{
\begin{tabular}{@{}l l@{}} 
\toprule 
\textbf{Notation} & \textbf{Description} \\
\midrule
$T_i$, $B_i$ & $i$-th task in a sequence $\{T_1, T_2, \dots, T_N\}$ and number of batches of $T_i$. \\
$D^s_i, D^t_i$ & Training/Testing datasets for task $T_i$: $D^s_i = \{ {\bf x}^i_j, {\bf y}^i_j \}^{n^i}_{j=1}$, $D^t_i = \{ {\bf x}^{t,i}_j, {\bf y}^{t,i}_j \}^{n^{t,i}}_{j=1}$. \\
$D^s_{i,b}, D^{buf}_{i,b}$ & $b$-th training/replay data batch in task $T_i$. \\
$n^i, n^{t,i}$ & Number of samples in $D^s_i$ and $D^t_i$. \\
$\mathbf{x}^i_j, \mathbf{y}^i_j$ & $j$-th training sample and label in $T_i$. \\
$\mathbf{x}^{t,i}_j, \mathbf{y}^{t,i}_j$ & $j$-th testing sample and label in $T_i$. \\
$\mathcal{X}, \mathcal{Y}$ & Space of data samples and labels: $\mathbf{x}^i_j \in \mathcal{X}$, $\mathbf{y}^i_j \in \mathcal{Y}$. \\
$\tilde{\Theta}, \theta^\star$ & Set of model parameters and the optimal set found via optimization. \\
$f_\theta(\cdot)$ & Classifier mapping $\mathcal{X}$ to $\mathcal{Y}$. \\
$\mathcal{L}(\cdot,\cdot)$ & Loss function, e.g., cross-entropy: $\mathcal{L}(\mathbf{y}, f_\theta(\mathbf{x}))$. \\
$H_2(\mathbf{X}^i), \hat{H}_2(\mathbf{X}^i)$ & Second-order Rényi entropy and its estimate for dataset $\mathbf{X}^i$. \\
$\hat{f}(\mathbf{x})$ & Estimated PDF using Gaussian kernel density estimator. \\
$V(\mathbf{X}^i)$ & Information Potential (IP) of dataset $\mathbf{X}^i$. \\
$D_{CS}(\mathbf{X}^i, \tilde{\mathbf{X}}^i)$ & Cauchy-Schwarz Divergence between $\mathbf{X}^i$ and $\tilde{\mathbf{X}}^i$. \\
$H(\mathbf{X}, \mathbf{Y})$ & Shannon entropy of the joint distribution of $\mathbf{X}$ and $\mathbf{Y}$. \\
$\hat{P}(\mathbf{X} | \mathbf{Y})$ & Estimated conditional probability distribution of $\mathbf{X}$ given $\mathbf{Y}$ using KDE. \\
$d_i^k$ & Average cosine distance for the $i$-th sample in class $\mathbf{Y}_k$. \\
$\mathcal{L}_{\text{info}}(\tilde{\mathbf{X}}^i)$ & Info-sample loss function for optimizing sample weights $\mathbf{w}$. \\
$r(\mathbf{w})$ & Regularizer in the info-sample loss function $\mathcal{L}_{\text{info}}$. \\
$\sigma$ & Bandwidth parameter for Gaussian kernel. \\
$\lambda_{H_2}, \lambda_{CS}$ & Trade-off parameters for Rényi entropy and CS Divergence in loss function. \\
$\lambda_{ksp}, \lambda_{L_1}, \lambda_H$ & Trade-off parameters for k-sparse, L1, and entropy regularization. \\
$w_j, {\hat w}_j$ & Selection weight for the $j$-th sample and its continuous relaxation variable. \\
${\mathcal{M}}^{\rm slow}_i, {\mathcal{M}}^{\rm fast}_i$ & Slow and fast memory buffers at the end of $T_i$. \\
$|{\mathcal{M}}_{buf}|$ & Capacity specified manually of memory buffer. \\

$f_{\rm category}({\mathcal{M}}^{\rm slow}_i, c )$ & Function returning samples in category $c$ from ${\mathcal{M}}^{\rm slow}_i$. \\
$f_d(\cdot,\cdot)$ & Function to calculate cosine distance between samples. \\
$\tilde{\mathbf{x}}^{\rm slow}_c$ & Central sample for category $c$ in ${\mathcal{M}}^{\rm slow}_i$. \\
$f_{\rm diversity}({\bf C}^{\rm slow}_c[j])$ & Diversity score of the $j$-th sample in category $c$. \\
$f_{\rm remove}({\bf C}^{\rm slow}_c[j])$ & Probability of removing the $j$-th sample in category $c$. \\

\bottomrule 
\end{tabular}
}
\label{notation_tab}
\end{table}

\subsection{Problem definition}
In continual learning, a model can not access the whole training dataset at one time and is learnt on a dynamically changing data stream. Let us define $\{T_1, T_2, \dots, T_N\}$ as a series of $N$ tasks, where each task ${T}_i$ contains a labelled training dataset $D^s_i = \{ {\bf x}^i_j, {\bf y}^i_j \}^{n^i}_{j=1}$ and a testing dataset  $D^t_i = \{ {\bf x}^{t,i}_j, {\bf y}^{t,i}_j \}^{n^{t,i}}_{j=1}$, where $n^i$ and $n^{t,i}$ denote the total number of samples for $D^s_i$ and $D^t_{i}$, respectively. ${\bf x}^{t,i}_j \in {\mathcal{X}}$ and ${\bf y}^{t,i}_j \in {\mathcal{Y}}$ represent the $j$ -th test data sample and the associated class label, respectively. We employ the superscript $s$ and $t$ for $D^s_i$ and $D^t_i$ to distinguish training and testing datasets, respectively. ${\mathcal{X}}$ and ${\mathcal{Y}}$ denote the image and class label space, respectively. The goal of a model in a continuous learning scenario is to find the optimal solution from a set of parameters $\tilde{\Theta}$, which can minimize the training loss in all tasks $\{T_1,\cdots,T_i\}$ in the $i$-th task learning (${T}_i$), expressed as~:

\begin{equation}
    \begin{aligned}
    \theta^\star = \underset{\theta \in \tilde{\Theta} }{\operatorname{argmin}} 
\frac{1}{i} \sum^i_{j=1} 
\sum^{n^j}_{j'=1}
{\mathcal{L}}
\left({\bf y}^j_{j'}, f_{\theta}({\bf x}^j_{j'} ) \right)\,,
\label{eq:problem}
    \end{aligned}
\end{equation}

\noindent where $\theta^\star$ is the set of optimal model parameters and $f_\theta(\cdot) \colon \mathcal{X} \to \mathcal{Y}$ is a classifier that takes ${\bf x}^j_{j'}$ as input and outputs the predicted label. ${\mathcal{L}}(\cdot,\cdot)$ is a loss function that can be implemented using the cross-entropy loss. However, searching for the optimal parameter set $\theta^\star$ using Eq.~\eqref{eq:problem} in continuous learning is intractable, since the model can only access the data samples of the current task learning ($T_i$) and all previous tasks $\{ T_1,\cdots,T_{i-1} \}$ are inaccessible. The continual learning studies aim to develop various technologies to find the optimal parameter set that can minimize the training loss on all tasks. Once the final task learning ($T_N$) is finished, we evaluate the model's performance on all testing datasets $\{ D^t_1,\cdots,D^t_N \}$.

\subsection{Information-Theoretic Dual Memory System}

Most existing memory-based methods usually consider managing a single memory system to preserve many critical data samples \cite{OnlineImbalanced}, which would not easily capture both long- and short-term knowledge during the whole learning procedure. In addition, using a single memory buffer would implement the memory optimization strategy by evaluating all memorized samples, leading to considerable computational costs. In this paper, we introduce a novel memory approach for continual learning, consisting of a fast memory buffer that aims to preserve recent information and a slow memory buffer that preserves long-term critical information about the data stream.

We implement the fast memory buffer optimization strategy by employing reservoir sampling for two main reasons: (1) it offers computational efficiency without incurring significant computational costs across various learning environments; (2) It allows for the random replacement of older memorized samples with newer ones, thereby ensuring the retention of current information over time; However, a key limitation of reservoir sampling lies in its ability to effectively store a diverse range of samples across all categories predominantly in a balanced continual learning scenario, in which the number of samples per category remains consistent across each task. This limitation is mitigated through the introduction of a slow memory buffer, which is designed to preserve critical data samples that encapsulate rich statistical information from all previously learned tasks. We achieve this by devising an innovative information-theoretic memory optimization approach that selectively retains a small subset of data samples from each task within the slow memory buffer, with the objective of maintaining diverse information across all previously encountered categories to the greatest extent possible. Prior to elucidating the proposed approach, we will first outline the technological framework underpinning the information theory methodology. Let us define ${\bf X}^i$ and ${\bf Y}^i$ as the random variables over the data and label space, associated with the $i$-th task. Furthermore, let $f({\bf x})$ represent the probability density function of \({\bf X}^i\).
According to the definition of second-order Rényi entropy, we estimate the sensitivity of the random variable ${\bf X}^i$ to high-probability events~:
\begin{equation}
H_2\left(\mathbf{X}^i\right) = \frac{1}{2} \log\big( \int_{D^s_i} f^2({\bf x}) \, d{\bf x} \big)\,.
\end{equation}

We assume that the random variable ${\bf X}^i$ follows the iid condition. The underlying probability density $f({\bf x})$ can be estimated using the Gaussian kernel density estimator, expressed as~:
\begin{equation}
\begin{aligned}
{f}({\bf x}) = \frac{1}{n^i} \sum_{n=1}^{n^i} k_{\sigma}({\bf x}, {\bf x}_n) = \frac{1}{n^i} \sum_{n=1}^{n^i} \exp\big(-\frac{\|{\bf x} - {\bf x}_n\|_2^2}{2\sigma^2}\big)\,,
\end{aligned}
\end{equation}

\noindent
where $\sigma$ is the bandwidth, and \(\|\cdot\|\) represents the \(L_2\) norm. To ensure high sensitivity to high-probability events during sampling, we consider estimating the second-order Rényi entropy of ${\bf X}^i$. Based on the information theory framework, the second-order Rényi entropy of the training dataset for the $i$-th task (${T}_i$) can be estimated as~:

\begin{equation}
\begin{aligned}
{H}_2\left(\mathbf{X}^i\right) = -\log\big(V\left(\mathbf{X}^i\right)\big) = -\log\Big(\frac{1}{(n^i)^2} \sum_{m=1}^{n^i} \sum_{n=1}^{n^i} k_{\sqrt{2}\sigma}\left({\bf x}_m, {\bf x}_n\right)\Big) \,,
\end{aligned}
\end{equation}

\noindent 
As a result, we can employ the second-order Rényi entropy to estimate the diversity of a subset $D^s_i[1:sn]$ from the training dataset $D^s_i$ of the $i$-th task ($T_i$), where $sn$ denotes the number of data samples for the subset. 

Let $\tilde{\mathbf{X}}^i$ represent the random variable defined over the space ${\mathcal{X}}$, associated with the subset $D^s_i[1:sn]$. Beyond the diversity of the samples, the statistical information gap between the subset $D^s_i[1:sn]$ and the original training dataset $D^s_i$ is pivotal for effective sample selection. An ideal subset $D^s_i[1:sn]$ should encapsulate abundant statistical information from the original training dataset $D^s_i$, ensuring that the model trained on $D^s_i[1:sn]$ retains robust performance. To accomplish this objective, we propose utilizing the Cauchy-Schwarz (CS) divergence to quantify the statistical information gap. Assuming that $\mathbf{X}^i$ and $\tilde{\mathbf{X}}^i$ satisfy the independent and identically distributed (iid) condition, we can compute the CS divergence between $\mathbf{X}^i$ and $\tilde{\mathbf{X}}^i$ as follows:

\begin{equation}
\begin{aligned}
D_{CS}\big(\mathbf{X}^i, \tilde{\mathbf{X}}^i\big) = 2H_2\big(\mathbf{X}^i, \tilde{\mathbf{X}}^i\big) - H_2\big(\mathbf{X}^i\big) - H_2\big(\tilde{\mathbf{X}}^i\big)\,,
\end{aligned}
\end{equation}
\noindent

where $H_2\big(\mathbf{X}^i,\tilde{\mathbf{X}}^i\big)$ denotes the second-order cross-Rényi entropy, which is estimated via the cross-information potential~:

\begin{equation}
\begin{aligned}
{H}_2\big(\mathbf{X}^i, \tilde{\mathbf{X}}^i\big) = -\log\big(V\big(\mathbf{X}^i, \tilde{\mathbf{X}}^i\big)\big) = -\log\Big(\frac{1}{n^i \cdot \tilde{n}^i} \sum_{m=1}^{n^i} \sum_{n=1}^{\tilde{n}^i} k_{\sqrt{2}\sigma}\left({\bf x}_m^i, \tilde{\bf x}_n^i\right)\Big) \,,
\label{eq:selection}
\end{aligned}
\end{equation}
where $\tilde{n}^i = sn$ signifies the sample count for the subset $D^s_i[1:sn]$. Eq~\eqref{eq:selection} serves as a criterion for selecting a suitable subset that maximally conveys the statistical information present in the training dataset $D^s_i$. In the subsequent section, we present an innovative memory optimization approach for the slow memory buffer, based on the principles of information theory.

\subsection{The Optimization Strategy for the Slow Memory Buffer}
In this section, we present a novel strategy for optimizing memory usage in the slow memory buffer. The objective is to choose a fixed number of samples from the dataset $D^s_i$, which can accurately represent its underlying data structure. To determine which samples should be incorporated into the slow memory buffer, we propose assigning a selection weight (represented as a binary variable) to each sample. A weight of "1" signifies that the corresponding sample has a high likelihood of being included in the slow memory buffer. To facilitate the selection of the most pertinent data samples, we first define an information cost function that assigns a high score to the most important samples, as expressed by the following equation~:
\begin{equation}
\begin{aligned}
J\big(\tilde{\mathbf{X}}^i\big) = \lambda_{H_2} \cdot H_2\big(\tilde{\mathbf{X}}^i\big) + \lambda_{CS} \cdot D_{CS}\big(\mathbf{X}^i, \tilde{\mathbf{X}}^i\big)\,,
\label{optimization1}
\end{aligned}
\end{equation}

\noindent
where \(\lambda_{H_2} \in \mathbb{R}^-\) and \(\lambda_{CS} \in \mathbb{R}^+\) are two hyperparameters controlling the importance of the second-order Renyi entropy term and  Cauchy-Schwar divergence term, respectively. The term $H_2\big(\tilde{\mathbf{X}}^i\big)$ in Eq.~\eqref{optimization1} aims to estimate the diversity of the data samples and is useful to address imbalanced data scenarios. The term $D_{CS}(\mathbf{X}^i, \tilde{\mathbf{X}}^i)$ in Eq.~\eqref{optimization1} is used to quantify the discrepancy between the selected subset and the real training dataset. Let $\bf w$ be a selection weight vector, where each $w_j \in (0,1)$ denotes the $j$-th dimension of ${\bf w}$ and is a selection probability for the $j$-th sample $\{{\bf x }^i_j, {\bf y}^i_j\}$ from the dataset ${D}^s_i$. The sample selection process can be formulated as an optimization procedure~:
\begin{equation}
\begin{aligned}
&\min {\mathcal{L}}_{\rm info}\big(\tilde{\mathbf{X}}^i\big)\,,
\\&
{\mathcal{L}}_{\rm info}\big(\tilde{\mathbf{X}}^i\big) = \lambda_{H_2}  H_2\left(\mathbf{X}^i \mathbf{w}\right) + \lambda_{CS}  D_{CS}\left(\mathbf{X}^i, \mathbf{X}^i \mathbf{w}\right) + r\left(\mathbf{w}\right)\,.
\label{optimization2}
\end{aligned}
\end{equation}
Compared to Eq.~\eqref{optimization1},  Eq.~\eqref{optimization2} involves a regularization term $r({\bf w})$ that is defined as~:
\begin{equation}
\begin{aligned}
r\left(\mathbf{w}\right) &= \lambda_{ksp} \big| sn - \sum_{j=0}^{n^i} w_j \big| + \lambda_{L_1} \sum_{i=0}^{n^i} \left| w_j \right| \\&- \lambda_H \sum_{j=0}^{n^i} \left( w_j \cdot \log w_j + \left(1 - w_j\right) \cdot \log\left(1 - w_j\right) \right)\,.
\label{regularization}
\end{aligned}
\end{equation}
We have found through empirical analysis that omitting the regularization term $r({\bf w})$ in Eq.~\eqref{optimization2} leads each sample weight to gravitate towards 1 throughout the optimization process. Consequently, this prevents us from selecting a suitable subset from the training dataset $D^s_i$. The L1 regularization term in Eq.~\eqref{regularization} is designed to manage the sparsity of sample weights, which facilitates an increase in the weights assigned to more informative and crucial samples while diminishing the weights of those deemed less significant. Furthermore, the entropy regularization term included in Eq.~\eqref{regularization} is intended to expedite the convergence of the optimization process outlined in Eq.~\eqref{optimization2}. We employ the gradient descent algorithm to optimize the sample weight vector ${\bf w}$, as articulated below~:
\begin{equation}
    \begin{aligned}
       {\bf w} = {\bf w} + l_{\rm w} \nabla_{\bf w}  {\mathcal{L}}_{\rm info}\big(\tilde{\mathbf{X}}^i\big) \,,
       \label{weightOptimization}
    \end{aligned}
\end{equation}
\noindent where $l_{\rm w}$ is a learning rate. Once the optimization procedure is converged, the weight of the more informative sample will approximate 1.

Nonetheless, the task of optimizing the sample weight vector $\bf w$ as described in Eq~\eqref{weightOptimization} presents significant computational challenges, primarily because each sample weight $w_j$ is a discrete variable that lacks differentiability in the context of the gradient descent method. To mitigate this limitation, we propose the development of a continuous and differentiable selection weight vector ${\hat {\bf w}}$, while also incorporating the use of the concrete distribution to produce differentiable discrete variables~:
\begin{equation}
\begin{aligned}
{w}_j &= \text{Bernoulli}({\hat w}_j) \\&= \sigma \big(\frac{1}{\lambda} \left(\log {\hat w}_j - \log (1 - {\hat w}_j) + \log u - \log(1 - u)\right)\big)\,,
\label{weightSampleing}
\end{aligned}
\end{equation}
\noindent
where $u \in (0,1)$ and ${w}_j $ is the continuous relaxation variable of ${\hat w}_j$. $\lambda$ is a temperature parameter for the concrete distribution. By using the sampling process defined in Eq.~\eqref{weightSampleing}, we propose to optimize the selection weight vector $\hat{{\bf w}}$ by~:
\begin{equation}
    \begin{aligned}
       {\hat{\bf w}} = {\hat{\bf w}}  + l_{\rm w} \nabla_{{\hat{\bf w}}}  {\mathcal{L}}_{\rm info}\big(\tilde{\mathbf{X}}^i\big) \,.
       \label{weightOptimization2}
    \end{aligned}
\end{equation}
The resulting selection weight vector ${\bf w}$ is used to guide storing data samples into the slow memory buffer.

\subsection{Memory Allocation Mechanism via A Balanced Sample Selection Approach}

To optimize the utilization of the proposed dual memory system, we ensure that both the fast and slow memory buffers are employed to retain as many data samples as feasible until the total of stored samples, represented as $|{\mathcal{M}}^{\rm slow}_i| + |{\mathcal{M}}^{\rm fast}_i|$, reaches the predetermined maximum memory capacity, $M^{\rm max}$. Here, ${\mathcal{M}}^{\rm slow}_i$ and ${\mathcal{M}}^{\rm fast}_i$ refer to the slow and fast memory components updated at the $i$-th task learning, respectively. The quantities $|{\mathcal{M}}^{\rm slow}_i|$ and $|{\mathcal{M}}^{\rm fast}_i|$ denote the total number of samples retained in each memory buffer. After each task switch, it is necessary for the slow memory buffer to purge a portion of the stored samples to create sufficient memory capacity for the integration of new data from the current task. Consequently, the slow memory buffer can effectively gather essential data samples across all tasks over time.

The central challenge associated with the removal of memorized samples from ${\mathcal{M}}^{\rm slow}_i$ lies in the fact that the slow memory buffer would eliminate varying quantities of samples across different categories, potentially resulting in a data imbalance issue. To overcome this challenge, we introduce an innovative Balanced Sample Selection (BSS) approach designed to appropriately remove samples from each category, thereby ensuring a more equitable distribution of memorized samples. Let $K_i$ denote the count of observed categories in ${\mathcal{M}}^{\rm slow}_i$ that is updated during the $i$-th task learning, and let $f_{\rm category}({\mathcal{M}}^{\rm slow}_i,c)$ represent a sample filtering function that yields a collection of memorized samples corresponding to the $c$-th category, articulated as follows~:
\begin{equation}
   \begin{aligned}
f_{\rm category}({\mathcal{M}}^{\rm slow}_i,c ) = \{ {\bf x}^{\rm slow}_j \,|\, f_{\rm label}({\bf x}^{\rm slow}_j) = c, j=1,\cdots,| {\mathcal{M}}^{\rm slow}_i| \} \,,
   \end{aligned} 
\end{equation}
\noindent where $f_{\rm label}(\cdot)$ is a function that returns the actual class label for a specified sample, and denotes ${\bf x}^{\rm slow}_j$ as the $j$-th memorized instance from the slow memory buffer ${\mathcal{M}}^{\rm slow}_i$. Additionally, let define ${\bf C}^{\rm slow}_c = f_{\rm category}({\mathcal{M}}^{\rm slow}_i, c)$ as the set of memorized instances pertinent to the $c$-th category. Our initial step involves identifying the central sample for the $c$-th category utilizing the samples in ${\bf C}^{\rm slow}_c$, formalized as follows~:
\begin{equation}
    \begin{aligned}
     {\tilde{\bf x}}^{\rm slow}_c  = \underset{ {\bf C}^{\rm slow}_c[j'] ,j'=1,\cdots, |{\bf C}^{\rm slow}_c| } {\operatorname{argmin}} \Big\{
     \frac{1}{|{\bf C}^{\rm slow}_c|-1}
     \sum^{|{\bf C}^{\rm slow}_c|}_{j=1, j \neq j' } \big\{  f_{d}( {\bf C}^{\rm slow}_c[j], {\bf C}^{\rm slow}_c[j'] ) \big\} \Big\} \,,
     \label{centralSample}
    \end{aligned}
\end{equation}
\noindent where ${\bf C}^{\rm slow}_c[j']$ denotes the $j'$-th sample from ${\bf C}^{\rm slow}_c$ and $|{\bf C}^{\rm slow}_c|$ represents the number of samples for ${\bf C}^{\rm slow}_c$. Eq.~\eqref{centralSample} aims to find the central sample $ {\tilde{\bf x}}^{\rm slow}_c$ that has the shortest distance with respect to other remaining samples of ${\bf C}^{\rm slow}_c$. In this paper, we implement $f_d(\cdot,\cdot)$ using the cosine distance and other distance measures will be investigated in our future study. The cosine distance is defined as~:
\begin{equation}
    \begin{aligned}
        f_d( {\bf x}_j, {\bf x}_{j'} ) =
          \frac{ \sum^{d}_{s=1} \{ {\bf x}_j[s]  {\bf x}_{j'}[s] \} }{ \sqrt{ \sum^{d}_{s=1} ({\bf x}_j[s])^2 }  \sqrt{ \sum^{d}_{s=1} ({\bf x}_{j‘}[s])^2 } } \,,
    \end{aligned}
\end{equation}
\noindent where ${\bf x}_j[s]$ and ${\bf x}_{j'}[s]$ denote the $s$-th dimension of the data sample ${\bf x}_j$ and ${\bf x}_{j'}$, respectively. $d$ is the data dimension. By using Eq.~\eqref{centralSample}, we can obtain the central samples $\{  {\tilde{\bf x}}^{\rm slow}_1,\cdots, {\tilde{\bf x}}^{\rm slow}_{K_i}  \}$ for all seen categories from ${\mathcal{M}}^{\rm slow}_i$ updated at the $i$-th task learning. To decide which sample should be removed from ${\mathcal{M}}^{\rm slow}_i$, we propose a novel sample diversity evaluation function, expressed as~:
\begin{equation}
    \begin{aligned}
        f_{\rm diversity} ({\bf C}^{\rm slow}_c[j] ) =  f_{d}( {\bf C}^{\rm slow}_c[j], {\tilde{\bf x}}^{\rm slow}_c ) \,.
        \label{diversity}
    \end{aligned}
\end{equation}
Eq.~\eqref{diversity} quantifies the distance between the $j$-th stored sample ${\bf C}^{\rm slow}_c[j]$ and the central sample ${\tilde{\bf x}}^{\rm slow}_c$, which serves as a diversity metric. A high value obtained from this equation suggests that ${\bf C}^{\rm slow}_c[j]$ possesses distinct semantic information relative to ${\tilde{\bf x}}^{\rm slow}_c$, warranting its retention in the slow memory buffer to enhance sample diversity. Conversely, a low diversity score implies the data is similar to the central sample and can be eliminated without significant loss of statistical information. 

Upon completion of the new task learning ($T_{i+1}$), it is essential to eliminate numerous memorized samples from ${\mathcal{M}}^{\rm slow}_i$ to accommodate new samples and mitigate the effects of forgetting. Let $K_{i+1}$ denote the number of categories encountered during the $(i+1)$-th task learning. An ideal slow memory buffer should sustain a consistent quantity of memorized samples across each category, which can be represented as $(M^{\rm max} / 2) / K_{i+1}$. For every collection of memorized samples in a category ${\bf C}^{\rm slow}_c, c=1,\cdots,K_i$, we compute the probability ${\mathcal{S}}_{c,j} = f_{\rm remove}( {\bf C}^{\rm slow}_c[j] )$ that each sample ${\bf C}^{\rm slow}_c[j]$ will be removed, as follows~:
\begin{equation}
    \begin{aligned}
        f_{\rm remove}( {\bf C}^{\rm slow}_c[j] ) = 1- \frac{ f_{\rm diversity} ({\bf C}^{\rm slow}_c[j] )}{ \sum^{|{\bf C}^{\rm slow}_c|}_{j'=1,j' \neq j} \big\{ f_{\rm diversity} ({\bf C}^{\rm slow}_c[j’] \big\} }  \,.
        \label{removepro}
    \end{aligned}
\end{equation}
Based on Eq.~\eqref{removepro}, we remove an appropriate number of memorized samples per category from the slow memory buffer by~:
\begin{equation}
    \begin{aligned}
       {\bf C}^{\rm slow}_c &= \{  {\bf C}^{\rm slow}_c[j] \,|\, f_{\rm remove}( {\bf C}^{\rm slow}_c[j] ) > f_{\rm remove}( {\bf C}^{\rm slow}_c[j+1] ), \\& j= 1,\cdots, (M^{\rm max} / 2) / K_{i+1} \} \,, c=1,\cdots,K_i \,.
       \label{sampleRemove}
    \end{aligned}
\end{equation}
By employing Eq~\eqref{sampleRemove}, each subset ${\bf C}^{\rm slow}_c$, where $c=1,\cdots,K_i$, maintains an equivalent number of retained samples. Furthermore, the slow memory buffer ${\mathcal{M}}^{\rm slow}_i$ possesses additional memory capacity, enabling it to accommodate a total of $\frac{M^{\rm max}}{2K_{i+1}}(K_{i+1} - K_{i})$ data samples for the subsequent task ($(i+1)$-th task). We update from ${\mathcal{M}}^{\rm slow}_i$ to ${\mathcal{M}}^{\rm slow}_{i+1}$ by selectively integrating data samples from the training dataset ${D}^s_{i+1}$ into the slow memory buffer.

\begin{figure}[t]
    \centering
    \includegraphics[trim=2.05cm 2.1cm 2.4cm 1.7cm, clip, width=1.0\textwidth, page=2]{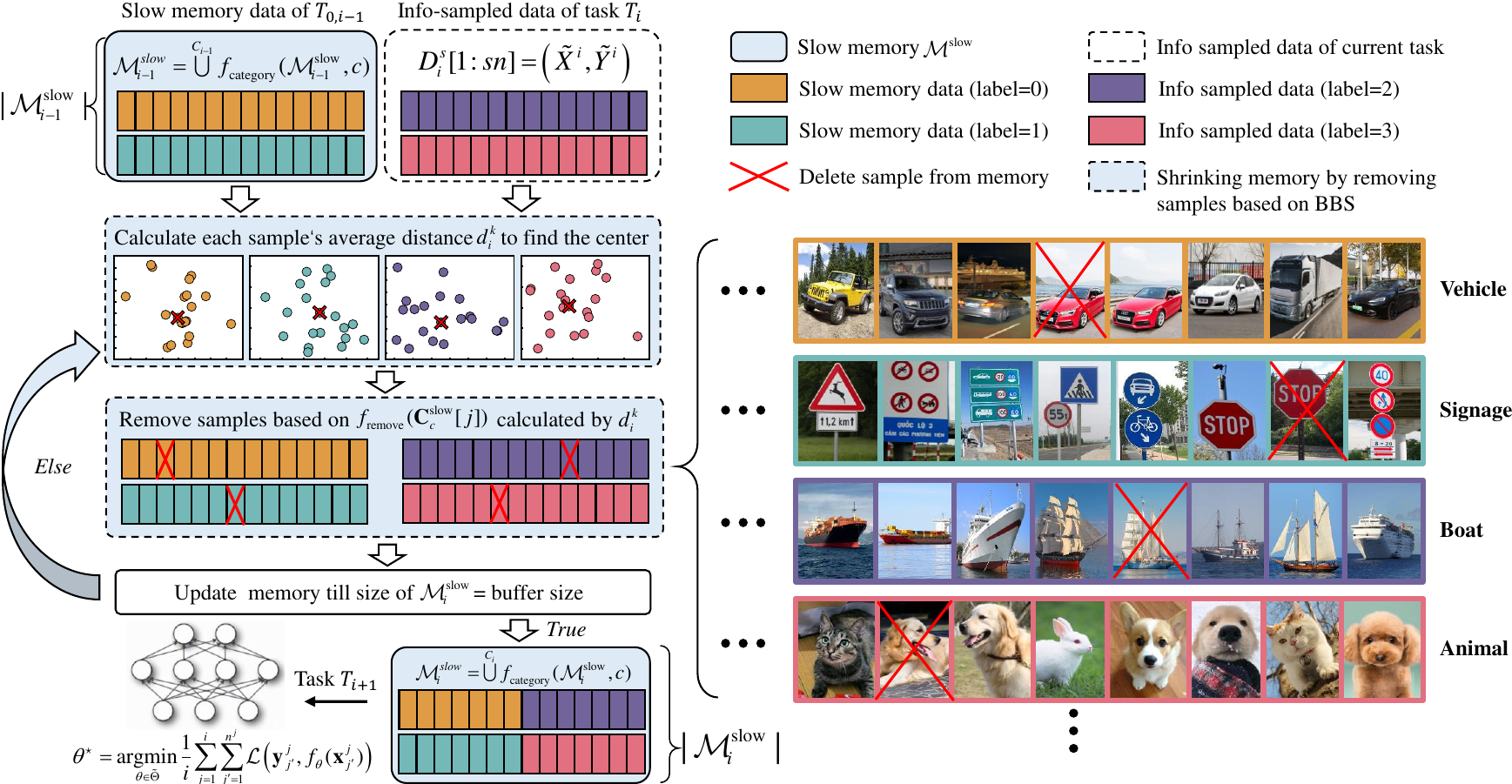}
    \caption{The optimization process of the proposed balanced sample selection approach. The first step is to define the central sample for each category for the slow memory buffer. The second step is to estimate the diversity score for each memorized sample from the slow memory buffer. Finally, we maintained a balanced label distribution for the samples in the slow memory while removing duplicate or similar sample, which have small average distances, thereby maximizing the diversity of the stored samples.}
\label{fig:informationApproach}
\end{figure}

\subsection{The Algorithm Framework} 

Given that the proposed dual memory architecture can seamlessly integrate with current continual learning techniques with minimal alterations, we propose to establish our memory system based on a widely recognized and fundamental baseline known as DER++ \cite{DER}. This baseline utilizes a singular memory buffer and employs reservoir sampling \cite{reservoir} for memory updates. In this study, we substitute the singular memory buffer of DER++ with the proposed dual memory framework, and we define the primary objective function for model training as follows~:
\begin{equation}
\begin{aligned}
{\mathcal{L}}_{\rm main} &=
\mathbb{E}_{({\bf x},{\bf y}) \sim {\rm P}_{D^s_i}} \big[ {\mathcal{L}}( {\bf y}, f_{\theta_i}({\bf x}))  \big] + \alpha  \mathbb{E}_{({\bf x}, {\bf z} )\sim {\rm P}_{{\mathcal{M}}^{\rm slow}_i \otimes {\mathcal{M}}^{\rm fast}_i}} [  || 
{\bf z} - f_{\theta_i}( {\bf x}) ||^2 ]
\\&+
\beta  \mathbb{E}_{({\bf x},{\bf y}, {\bf z} )\sim {\rm P}_{{\mathcal{M}}^{\rm slow}_i \otimes {\mathcal{M}}^{\rm fast}_i}} [  {\mathcal{L}}({\bf y}, f_{\theta_i}( {\bf x}) ]
\,,
\label{der_eq3}
\end{aligned}
\end{equation}
\noindent where ${\rm P}_{D^s_i}$ and ${\rm P}_{{\mathcal{M}}^{\rm slow}_i \otimes {\mathcal{M}}^{\rm fast}_i}$ denote the distribution of the dataset ${D}^s_i$ and the combined memory buffer ${\mathcal{M}}^{\rm slow}_i \otimes {\mathcal{M}}^{\rm fast}_i$ where $\otimes$ combines two datasets. $\lambda'$ is a hyperparameter to control the importance of the updating process on memorized samples. We provide the pseudocode and the detailed learning procedure for the proposed dual memory system in \textbf{Algorithm 1} and Fig.~\ref{fig:training}, which consists of three steps~:

\vspace{3pt}
\noindent \textBF{Step 1. Update the fast memory buffer~:} We continually update the fast memory buffer using the reservoir sampling.

\vspace{3pt}
\noindent \textBF{Step 2. Remove memorized samples~:} Once the current task learning ($T_i, i>1$) is finished, we first determine the central sample for each category in the slow memory buffer using Eq.~\eqref{centralSample}. Then we design a function (Eq.~\eqref{removepro}) to calculate the removing probability for each memorized sample. We remove memorized samples from the slow memory buffer using Eq.~\eqref{sampleRemove} to ensure the category balance. 

\vspace{3pt}
\noindent \textBF{Step 3. Add new data samples~:} We optimize the selection weight probability vector using Eq.~\eqref{weightOptimization2}. We selectively store new data samples with high selection probabilities in the slow memory buffer.

\begin{algorithm}[t]
\caption{The training algorithm for the proposed dual memory system}
\label{alg:Information-Theoretic Dual Memory System}
\begin{algorithmic}[1]
\setstretch{1.15}
\REQUIRE The tasks $\{T_1, T_2, \dots, T_N\}$, parameters $\theta$, scalars $\alpha$ and $\beta$, learning rate $\lambda$
\STATE ${\mathcal{M}}^{\rm slow} \leftarrow \{\}$, ${\mathcal{M}}^{\rm fast} \leftarrow \{\}$
\FOR{$i$ in $N$}
    \FOR{each $({\bf x}, {\bf y})$ in $D^s_i$}
    \STATE \textBF{The first step~: Update the fast memory buffer}
        \STATE ${\mathcal{M}}^{\rm fast}_i \leftarrow \text{reservoir}({\mathcal{M}}^{\rm fast}_i, ({\bf x}, {\bf z}, {\bf y}))$
        \STATE $({\bf x}'_{f}, {\bf z}'_{f}, {\bf y}'_{f}), ({\bf x}'_{s}, {\bf z}'_{s}, {\bf y}'_{s}) \leftarrow \text{sample}({\mathcal{M}}^{\rm fast}_i, {\mathcal{M}}^{\rm slow}_i)$
        \STATE $({\bf x}', {\bf z}', {\bf y}') \leftarrow \text{concat}(({\bf x}'_{f}, {\bf z}'_{f}, {\bf y}'_{f}), ({\bf x}'_{s}, {\bf z}'_{s}, {\bf y}'_{s}))$
        \STATE $({\bf x}''_{f}, {\bf z}''_{f}, {\bf y}''_{f}), ({\bf x}''_{s}, {\bf z}''_{s}, {\bf y}''_{s}) \leftarrow \text{sample}({\mathcal{M}}^{\rm fast}_i, {\mathcal{M}}^{\rm slow}_i)$
        \STATE $({\bf x}'', {\bf z}'', {\bf y}'') \leftarrow \text{concat}(({\bf x}''_{f}, {\bf z}''_{f}, {\bf y}''_{f}), ({\bf x}''_{s}, {\bf z}''_{s}, {\bf y}''_{s}))$
        \STATE ${\bf x}_t, {\bf x}_t', {\bf x}_t'' \leftarrow \text{augment}({\bf x}, {\bf x}', {\bf x}'')$
        \STATE ${\bf z} \leftarrow h_{\theta}({\bf x}_t)$
        \STATE $reg \leftarrow \alpha \|{\bf z}' - h_{\theta}({\bf x}_t')\|_2^2 + \beta {\mathcal{L}}({\bf y}'', f_{\theta}({\bf x}_t''))$
        \STATE $\theta \leftarrow \theta +  \cdot \nabla_{\theta} [\ell({\bf y}, f_{\theta}({\bf x}_t)) + reg]$
    \ENDFOR
    \IF{ $(i + 1 < N)$ and $|{\mathcal{M}}^{\rm slow}|$}
   \STATE \textBF{The second step~: Remove memorized samples}
\FOR {$c=0,c<K_i$}
\STATE Remove samples from ${\bf{C}}^{\rm slow}_c$ using Eq.~\eqref{sampleRemove}
\ENDFOR
   \STATE \textBF{The third step~: Add new data samples}
\STATE Perform \textbf{Algorithm 2} to add samples from the new dataset $D^s_{i+1}$ into ${\mathcal{M}}^{\rm slow}_i$
    \STATE ${\mathcal{M}}^{\rm fast} \leftarrow \{\}$
    \ENDIF
\ENDFOR
\end{algorithmic}
\end{algorithm}

\begin{algorithm}[t]
\caption{An information-theoretic memory optimization approach }
\label{alg:example2}
\begin{algorithmic}[1]
\setstretch{1.15}
\REQUIRE The dataset $D^s_{i+1}$ of the $i+1$-th task, sample weight $\mathbf{\hat w}$, sample learning rate $l_{\rm w}$, sample number \textit{sn}, parameter $\lambda_{ksp},\ \lambda_{L_1},\ \lambda_H$, slow memory ${\mathcal{M}}^{\rm slow}$.
\FOR{epoch in epochs}
\STATE $\mathbf{w} \leftarrow \text{Bernoulli}(\mathbf{\hat w})$
\STATE $\lambda_{H_2} \cdot H_2\big(\mathbf{X}^{i+1}\mathbf{w}\big) + \lambda_{CS} \cdot D_{CS}\big(\mathbf{X}^{i+1}, {\hat{\mathbf{X}}}^{i+1}\mathbf{w}\big)$
\STATE $\lambda_{ksp} \big| sn - \sum\limits_{j=0}^{n^{i+1}} w_j \big| + \lambda_{L_1} \sum\limits_{i=0}^{n^{i+1}} \left| w_j \right| - \lambda_H \sum\limits_{j=0}^{n^{i+1}} \left( w_j \cdot \log w_j + \left(1 - w_j\right) \cdot \log\left(1 - w_j\right) \right)$
\STATE ${\hat{\bf w}} = {\hat{\bf w}}  + l_{\rm w} \nabla_{{\hat{\bf w}}}  {\mathcal{L}}_{\rm info}\big(\tilde{\mathbf{X}}^i\big) $ 
\ENDFOR
\FOR{($j= 0, j <n^{i+1}$) }
\STATE ${w}_j = \sigma \big(\frac{1}{\lambda} \left(\log {\hat w}_j - \log (1 - {\hat w}_j) + \log u - \log(1 - u)\right)\big)$
\ENDFOR
\STATE Add data samples into ${\mathcal{M}}^{\rm slow}_{i+1}$ according to $\bf w$
\RETURN $\mathcal{M}^{\rm slow}_{i+1}$
\end{algorithmic}
\end{algorithm}

\section{Experiment}
\label{experiment}

We conduct experiments in continual learning, focusing on three principal scenarios~: Task Incremental Learning (Task-IL), Class Incremental Learning (Class-IL), and Domain Incremental Learning (Domain-IL).

In the Task-IL scenario, each training task operates with an independent label space. Conversely, in the Class-IL scenario, the tasks share a common label space. During evaluation, the model in the Task-IL framework receives the label space corresponding to the current task, while in the Class-IL approach, the model remains unaware of the specific task to which the sample belongs. In practice, datasets such as CIFAR-10 and Tiny ImageNet are segmented into 5 and 10 tasks, respectively, introducing 2 and 20 new classes per task in a consistent and fixed sequence across various iterations.

Domain-IL entails learning across multiple domains that share the same label space, yet exhibit different data distributions. The central challenge in these scenarios is ensuring that the model continues to learn while retaining previously acquired knowledge, amidst the ongoing evolution of data distributions. For this purpose, we employ two established protocols derived from the MNIST dataset: Permuted MNIST, which involves random pixel shuffling, and Rotated MNIST, which rotates images by a random angle within $[0, \pi)$. The total number of tasks in both protocols is 20.

\subsection{Evaluation Framework}

\textbf{Architecture:} In the context of the MNIST configurations, a multilayer perceptron (MLP) consisting of two hidden layers, each with 400 neurons, along with 2,000 memory slots, is implemented for the balanced scenario. Conversely, for the imbalanced scenario, a configuration featuring 100 neurons and a memory capacity of 300 is adopted. Regarding the CIFAR configurations, a streamlined variant of ResNet18 is employed, which is not pre-trained.

\textbf{Dataset:} Balanced data streams feature approximately equal data quantities per task. We examine five key benchmarks:

\begin{enumerate}
    \item \textbf{Split-MNIST:} Utilizes the MNIST dataset with its 60,000 training samples divided into 5 tasks, each representing pairs of consecutive digits.
    \item \textbf{Split-CIFAR10:} Divides the CIFAR-10 dataset into 5 tasks, each containing 2 labels and 10,000 training samples.
    \item \textbf{Sequential-Tiny ImageNet:} Involves the Tiny ImageNet dataset, partitioned into 20 tasks with 2,500 samples each from a pool of 50,000 training samples.
    \item \textbf{Rot-MNIST:} Rotates the entire MNIST dataset, resulting in 20 tasks, each with 60,000 training samples and 10,000 test samples, with each task containing a total of 10 label classes representing digits 0-9.
    \item \textbf{Perm-MNIST:} Involves pixel permutation of the entire MNIST dataset, generating a total of 20 tasks using different permutation vectors, where each task is a permuted version of the original image. Each task includes 60,000 training samples and 10,000 test samples, with each task containing a total of 10 label classes representing digits 0-9.
\end{enumerate}

For all these datasets, the evaluation process utilizes the entire original test subset, noted as \(S_{eval}\).

Imbalanced data streams present a more realistic scenario by eliminating the equality assumption regarding task duration. This experiment, which is less common and often requires artificial balancing, is used in both imbalanced Split-MNIST and two new challenging benchmarks. We consider using two imbalanced datasets, Imbalanced Split-MNIST and Imbalanced Split-CIFAR10. In the imbalanced learning setting, the first task ${T_1}$ contains 4,000 samples, while the sample size for the remaining tasks is 1/10 of that amount. 


\textbf{Augmentation:} For fairness in comparison, we do not apply data augmentation or transformation to the MNIST-based datasets. However, for more complex datasets such as CIFAR10 and Tiny ImageNet, we apply data augmentation techniques, including random cropping and horizontal flipping. This choice will also be applied to the competing models. Samples stored in the buffer do not undergo any data augmentation or transformation.

\textbf{Hyperparameter Selection:} We determine the hyperparameters by using Bayesian optimization based on distributed hyperparameter optimization on the validation set. Compared to random search and grid search, this method uses Bayesian statistics to iteratively select hyperparameters, guiding the next search based on previous results, leading to higher search efficiency and better convergence to optimal solutions.

\textbf{Training Process:} To ensure consistent comparison across different continual learning models, we use SGD as the optimizer for all models. For MNIST-based settings, we set the number of epochs to 1, which has been shown to be sufficient for model training. For the more complex CIFAR10 and Sequential Tiny ImageNet datasets, we follow the settings from previous work, with 50 and 100 epochs respectively. The batch size and replay batch size for existing continual learning models are set according to the optimal results reported in the literature, while our model parameters are consistent with those of DER++.

\begin{table*}[t]
    \centering
    \fontsize{9}{11}\selectfont
    \setlength{\tabcolsep}{1mm}{
    \begin{tabular}{@{}c l c c c c c c c@{}}
    \toprule
    \multirow{2}{*}{\textbf{Buffer}} & \multirow{2}{*}{\textbf{Method}} & \multicolumn{2}{c}{\textbf{Split-CIFAR10}} & \multicolumn{2}{c}{\textbf{Sequential-Tiny ImageNet}} & \multicolumn{1}{c}{\textbf{P-MNIST}} & \multicolumn{1}{c}{\textbf{R-MNIST}} \\
    \cmidrule(l){3-4}\cmidrule(l){5-6}\cmidrule(l){7-7}\cmidrule(l){8-8}
    & & \textbf{Class-IL} & \textbf{Task-IL} & \textbf{Class-IL} & \textbf{Task-IL} & \textbf{Domain-IL} & \textbf{Domain-IL} \\
    \midrule
    \multirow{6}{*}{\centering -} 
    & JOINT & 92.20{\scriptsize$\pm$0.15} & 98.31{\scriptsize$\pm$0.12} & 59.99{\scriptsize$\pm$0.19} & 82.04{\scriptsize$\pm$0.10} & 94.33{\scriptsize$\pm$0.17} & 95.76{\scriptsize$\pm$0.04} \\
    & SGD & \textBF{19.62{\scriptsize$\pm$0.05}} & 61.02{\scriptsize$\pm$3.33} & \textBF{7.92{\scriptsize$\pm$0.26}} & 18.31{\scriptsize$\pm$0.68} & 40.70{\scriptsize$\pm$2.33} & 67.66{\scriptsize$\pm$8.53} \\
    & EWC & 19.49{\scriptsize$\pm$0.12} & \textBF{68.29{\scriptsize$\pm$3.92}} & 7.58{\scriptsize$\pm$0.10} & 19.20{\scriptsize$\pm$0.31} & \textBF{75.79{\scriptsize$\pm$2.25}} & \textBF{77.35{\scriptsize$\pm$5.77}} \\
    & SI & 19.48{\scriptsize$\pm$0.17} & 68.05{\scriptsize$\pm$5.91} & 6.58{\scriptsize$\pm$0.31} & 36.32{\scriptsize$\pm$0.13} & 65.86{\scriptsize$\pm$1.57} & 71.91{\scriptsize$\pm$5.83} \\
    & LwF & 19.61{\scriptsize$\pm$0.05} & 63.29{\scriptsize$\pm$2.35} & 8.46{\scriptsize$\pm$0.22} & 15.85{\scriptsize$\pm$0.58} & - & - \\
    & PNN & - & 95.13{\scriptsize$\pm$0.72} & - & \textBF{67.84{\scriptsize$\pm$0.29}} & - & - \\
    \midrule
    \multirow{10}{*}{\centering \textbf{200}} 
    & ER & 44.79{\scriptsize$\pm$1.86} & 91.19{\scriptsize$\pm$0.94} & 8.49{\scriptsize$\pm$0.16} & 38.17{\scriptsize$\pm$2.00} & 72.37{\scriptsize$\pm$0.87} & 85.01{\scriptsize$\pm$1.90} \\
    & GEM & 25.54{\scriptsize$\pm$0.76} & 90.44{\scriptsize$\pm$0.94} & - & - & 66.93{\scriptsize$\pm$1.25} & 80.80{\scriptsize$\pm$1.15} \\
    & A-GEM & 20.04{\scriptsize$\pm$0.34} & 83.88{\scriptsize$\pm$1.49} & 8.07{\scriptsize$\pm$0.08} & 22.77{\scriptsize$\pm$0.03} & 66.42{\scriptsize$\pm$4.00} & 81.91{\scriptsize$\pm$0.76} \\
    & iCaRL & 49.02{\scriptsize$\pm$3.20} & 88.99{\scriptsize$\pm$2.13} & 7.53{\scriptsize$\pm$0.79} & 28.19{\scriptsize$\pm$1.47} & - & - \\
    & FDR & 30.91{\scriptsize$\pm$2.74} & 91.01{\scriptsize$\pm$0.68} & 8.70{\scriptsize$\pm$0.19} & 40.36{\scriptsize$\pm$0.68} & 74.77{\scriptsize$\pm$0.83} & 85.22{\scriptsize$\pm$3.35} \\
    & GSS & 39.07{\scriptsize$\pm$5.59} & 88.80{\scriptsize$\pm$2.89} & - & - & 63.72{\scriptsize$\pm$0.70} & 79.50{\scriptsize$\pm$0.41} \\
    & HAL & 32.36{\scriptsize$\pm$2.70} & 82.51{\scriptsize$\pm$3.20} & - & - & 74.15{\scriptsize$\pm$1.65} & 84.02{\scriptsize$\pm$0.98} \\
    & DER & 61.93{\scriptsize$\pm$1.79} & 91.40{\scriptsize$\pm$0.92} & 11.87{\scriptsize$\pm$0.78} & 40.22{\scriptsize$\pm$0.67} & 81.74{\scriptsize$\pm$1.07} & 90.04{\scriptsize$\pm$2.61} \\
    & DER++ & 64.88{\scriptsize$\pm$1.17} & \textBF{91.92{\scriptsize$\pm$0.60}} & 10.96{\scriptsize$\pm$1.17} & 40.87{\scriptsize$\pm$1.16} & 83.58{\scriptsize$\pm$0.59} & 90.43{\scriptsize$\pm$1.87} \\
    & \textBF{ITDMS\scriptsize(ours)} & \textBF{66.13{\scriptsize$\pm$1.46}} & 91.01{\scriptsize$\pm$0.63} & \textBF{15.84{\scriptsize$\pm$0.72}} & \textBF{43.92{\scriptsize$\pm$0.29}} & \textBF{85.01{\scriptsize$\pm$0.76}} & \textBF{91.53{\scriptsize$\pm$1.14}} \\
    \midrule
    \multirow{10}{*}{\centering \textbf{500}} 
    & ER & 57.74{\scriptsize$\pm$0.27} & 93.61{\scriptsize$\pm$0.27} & 9.99{\scriptsize$\pm$0.29} & 48.64{\scriptsize$\pm$0.46} & 80.60{\scriptsize$\pm$0.86} & 88.91{\scriptsize$\pm$1.44} \\
    & GEM & 26.20{\scriptsize$\pm$1.26} & 92.16{\scriptsize$\pm$0.69} & - & - & 76.88{\scriptsize$\pm$0.52} & 81.15{\scriptsize$\pm$1.98} \\
    & A-GEM & 22.67{\scriptsize$\pm$0.57} & 89.48{\scriptsize$\pm$1.45} & 8.06{\scriptsize$\pm$0.04} & 25.33{\scriptsize$\pm$0.49} & 67.56{\scriptsize$\pm$1.28} & 80.31{\scriptsize$\pm$6.29} \\
    & iCaRL & 47.55{\scriptsize$\pm$3.95} & 88.22{\scriptsize$\pm$2.62} & 9.38{\scriptsize$\pm$1.53} & 31.55{\scriptsize$\pm$3.27} & - & - \\
    & FDR & 28.71{\scriptsize$\pm$3.23} & 93.29{\scriptsize$\pm$0.59} & 10.54{\scriptsize$\pm$0.21} & 49.88{\scriptsize$\pm$0.71} & 83.18{\scriptsize$\pm$0.53} & 89.67{\scriptsize$\pm$1.63} \\
    & GSS & 49.73{\scriptsize$\pm$4.78} & 91.02{\scriptsize$\pm$1.57} & - & - & 76.00{\scriptsize$\pm$0.87} & 81.58{\scriptsize$\pm$0.58} \\
    & HAL & 41.79{\scriptsize$\pm$4.46} & 84.54{\scriptsize$\pm$2.36} & - & - & 80.13{\scriptsize$\pm$0.49} & 85.00{\scriptsize$\pm$0.96} \\
    & DER & 70.51{\scriptsize$\pm$1.67} & 93.40{\scriptsize$\pm$0.39} & 17.75{\scriptsize$\pm$1.14} & 51.78{\scriptsize$\pm$0.88} & 87.29{\scriptsize$\pm$0.46} & 92.24{\scriptsize$\pm$1.12} \\
    & DER++ & 72.70{\scriptsize$\pm$1.36} & \textBF{93.88{\scriptsize$\pm$0.50}} & 19.38{\scriptsize$\pm$1.41} & 51.91{\scriptsize$\pm$0.68} & 88.21{\scriptsize$\pm$0.39} & 92.77{\scriptsize$\pm$1.05} \\
    & \textBF{ITDMS\scriptsize(ours)} & \textBF{74.44{\scriptsize$\pm$0.40}} & 92.98{\scriptsize$\pm$0.73} & \textBF{21.38{$\pm$0.3}} & \textBF{52.20{\scriptsize$\pm$0.75}} & \textBF{88.64{\scriptsize$\pm$0.32}} & \textBF{93.47{\scriptsize$\pm$0.78}} \\
    \midrule
    \multirow{9}{*}{\centering \textbf{1000}} 
    & ER & 70.44{\scriptsize$\pm$0.55} & \textBF{95.34{\scriptsize$\pm$0.16}} & 12.85{\scriptsize$\pm$0.47} & 55.92{\scriptsize$\pm$0.90} & 84.61{\scriptsize$\pm$0.99} & 90.42{\scriptsize$\pm$1.07} \\
    & GEM & 22.22{\scriptsize$\pm$0.87} & 93.67{\scriptsize$\pm$0.32} & - & - & - & - \\
    & A-GEM & 20.16{\scriptsize$\pm$0.41} & 85.61{\scriptsize$\pm$2.01} & 7.98{\scriptsize$\pm$0.17} & 24.29{\scriptsize$\pm$1.28} & 72.48{\scriptsize$\pm$1.97} & 81.30{\scriptsize$\pm$5.33} \\
    & iCaRL & 67.27{\scriptsize$\pm$0.63} & 91.4{\scriptsize$\pm$1.06} & 29.24{\scriptsize$\pm$0.24} & 63.87{\scriptsize$\pm$0.25} & - & - \\
    & FDR & 23.62{\scriptsize$\pm$2.88} & 94.02{\scriptsize$\pm$0.64} & 13.54{\scriptsize$\pm$0.61} & 56.05{\scriptsize$\pm$0.71} & 86.89{\scriptsize$\pm$0.23} & 91.68{\scriptsize$\pm$1.01} \\
    & GSS & 53.53{\scriptsize$\pm$4.55} & 91.79{\scriptsize$\pm$2.16} & - & - & 75.58{\scriptsize$\pm$2.16} & 82.25{\scriptsize$\pm$2.42} \\
    & HAL & 49.97{\scriptsize$\pm$3.01} & 87.33{\scriptsize$\pm$1.46} & - & - & 83.35{\scriptsize$\pm$0.41} & 89.33{\scriptsize$\pm$2.01} \\
    & DER++ & 77.82{\scriptsize$\pm$0.81} & 94.99{\scriptsize$\pm$0.26} & 25.01{\scriptsize$\pm$0.89} & 58.05{\scriptsize$\pm$0.52} & 90.22{\scriptsize$\pm$0.18} & 93.82{\scriptsize$\pm$0.39} \\
    & \textBF{ITDMS} & \textBF{79.58{\scriptsize$\pm$0.75}} & 93.98{\scriptsize$\pm$0.55} & \textBF{26.77{\scriptsize$\pm$0.58}} & 57.55{\scriptsize$\pm$0.82} & 89.02{\scriptsize$\pm$0.49} & \textBF{94.59{\scriptsize$\pm$0.51}} \\
    \bottomrule
    \end{tabular}
    }
    \caption{The average classification of various models on four datasets.  }
    \label{tab:averAccuracy}
\end{table*}
\subsection{Experiment Results on the Balanced Data Stream}
\label{sec:balacnedResults}

In this section, we assess our model's efficacy utilizing balanced datasets and compare it against several established methodologies, including regularization-based techniques (such as oEWC and SI), knowledge distillation methods (e.g., iCaRL and LwF), a structure-oriented approach (PNN), and a range of rehearsal-based strategies (including ER, GEM, A-GEM, GSS, FDR, and HAL). For a thorough comparison, we implement a singular classifier trained with Stochastic Gradient Descent (SGD) across the entire training dataset, referred to as JOINT, which evades the pitfalls of network forgetting. This experiment setting enables us to identify both absolute and relative performance variances. We present the mean classification accuracy of the various models in Table.~\ref{tab:averAccuracy}.

\begin{figure}[h!]
    \centering
    \includegraphics[trim=6.1cm 0cm 6.9cm 0cm, clip, width=1.0\textwidth, page=7]{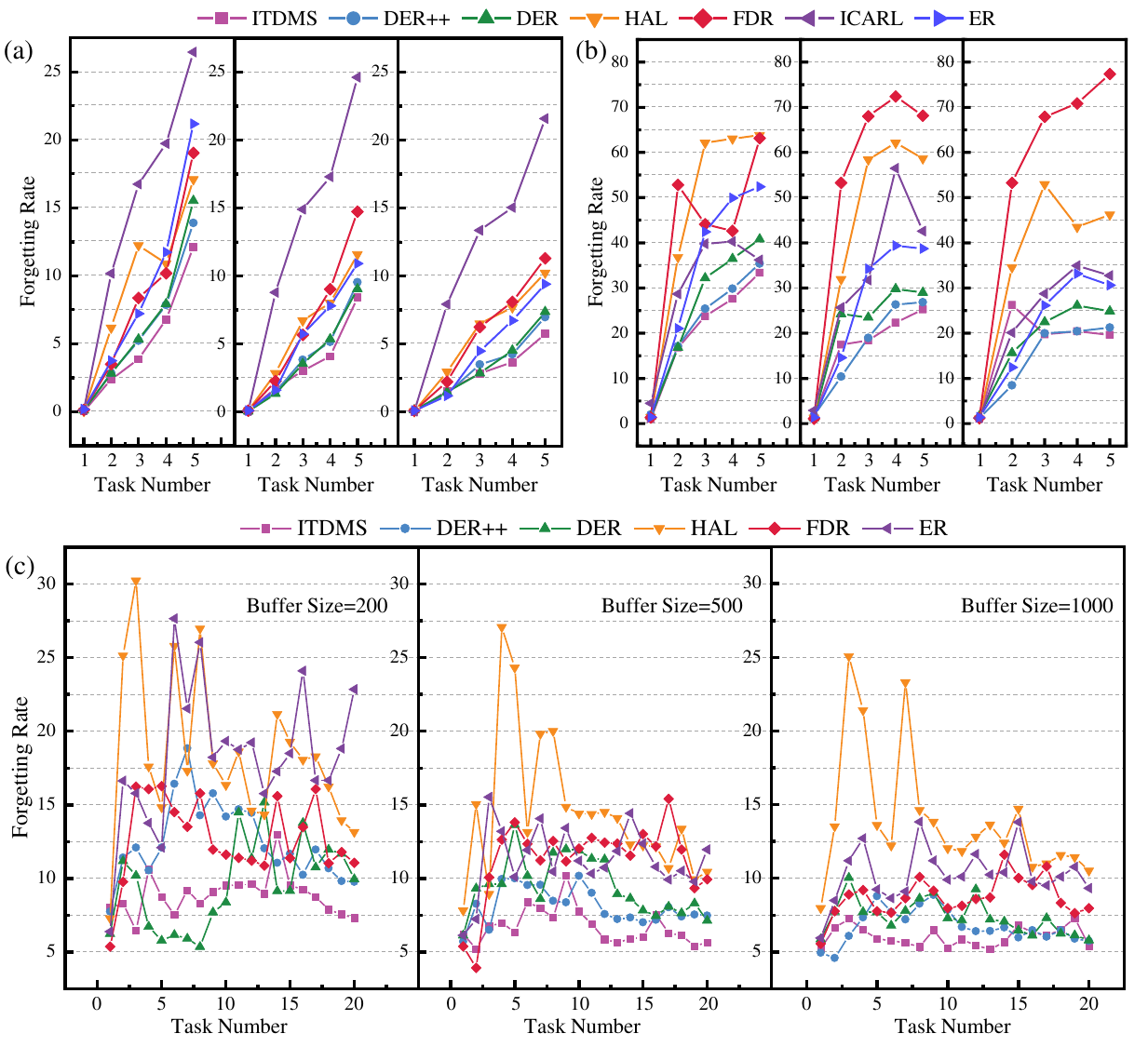}
    \caption{The forgetting analysis of various models. (a), (b), (c) report the results on the Split-MNIST, Split-CIFAR10 and R-MNIST, respectively. Each subfigure corresponding to a dataset is set with three different storage parameters: 200, 500, and 1000, from left to right.}
    \label{fig:Forgetting}
\end{figure}

The forgetting curves for various models with different memory configurations on the Split-MNIST, Split-CIFAR10, and R-MNIST datasets are illustrated in Figure~\ref{fig:Forgetting}a, Figure~\ref{fig:Forgetting}b, and Figure~\ref{fig:Forgetting}c, respectively. The results reveal that the proposed ITDMS consistently demonstrates the lowest rates of forgetting, significantly surpassing other methodologies. This underscores ITDMS's capacity to preserve acquired knowledge as the number of tasks escalates, showcasing remarkable stability and resilience against forgetting. Notably, even with the use of small buffer size, our model sustains outstanding performance, further validating its adaptability and efficacy across diverse continual learning settings.

The results presented in Table.~\ref{tab:averAccuracy} indicate that the proposed ITDMS offers a distinct superiority over all baseline models across the majority of continual learning contexts. By integrating our proposed memory strategy with the baseline (DER++), these outcomes illustrate that the performance improvements are attributable to our memory framework. Notably, in a more challenging continual learning scenario where memory capacity is severely restricted, the proposed memory system demonstrates a substantial enhancement in performance.

Previous continual learning technologies such as oEWC, SI, and LwF struggle with extended task sequences. Regularization-based approaches often degrade as task relationships evolve, while LwF's emphasis on preserving previous outputs leads to significant forgetting when tasks overlap minimally. PNN, a dynamic expansion model, despite its strengths in isolated tasks, suffers from scalability and memory issues in longer sequences. Memory-based methods such GEM, A-GEM, and GSS also falter, particularly with limited buffer sizes, where they fail to capture crucial sample information. In contrast, the proposed ITDMS approach can selectively store more important and informative data samples into the memory buffer, which leads to superior performance even in challenging continual learning settings.

Previous memory-based approaches such as GEM, A-GEM, and GSS are also impacted by performance deterioration, especially in contexts where buffer sizes are limited. These methods utilize gradient projection to mitigate the risk of forgetting; however, their efficacy is frequently undermined by task complexity and the accuracy demanded in gradient assessments. When faced with restricted buffer capacities, these techniques find it challenging to seize and preserve essential sample data, leading to significant declines in performance when learning new tasks. Notably, A-GEM, due to its streamlined update mechanisms, tends to exhibit particular instability in environments characterized by high task complexity.

\begin{table}[h]
\centering
    \fontsize{9}{11}\selectfont
\begin{tabular}{c ccc ccc ccc} 
\toprule
\textbf{Data set} & \multicolumn{3}{c}{\textbf{Split-MNIST}} & \multicolumn{3}{c}{\textbf{Split-CIFAR10}} & \multicolumn{3}{c}{\textbf{Tiny-ImageNet}} \\ 
\cmidrule(r){2-4} \cmidrule(r){5-7} \cmidrule(r){8-10} 
\textbf{Memory size} & 200 & 500 & 1000 & 200 & 500 & 1000 & 200 & 500 & 1000 \\ 
\midrule
\textbf{$\lambda_{H_2}$} & -1 & -1 & -1 & -1 & -1 & -1 & -1 & -1 & -1 \\ 
\textbf{$\lambda_{CS}$} & 1 & 1 & 10 & 1 & 1 & 1 & 10 & 10 & 100 \\ 
\textbf{$|{\mathcal{M}}^{\rm slow}|$} & 200 & 500 & 1000 & 200 & 500 & 1000 & 200 & 500 & 500 \\ 
\textbf{$|{\mathcal{M}}^{\rm fast}|$} & None & None & None & 200 & 500 & 1000 & 200 & 500 & 500 \\ 
\textbf{${\mathcal{D}}^{\rm slow}_b$} & 1.0 & 1.0 & 1.0 & 1.0 & 1.0 & 1.0 & 1.0 & 0.5 & 0.5 \\ 
\textbf{${\mathcal{D}}^{\rm fast}_b$} & 0 & 0 & 0 & 0.05 & 0.05 & 0.05 & 0.05 & 0.5 & 0.5 \\ 
\textbf{Reset ${\mathcal{M}}^{\rm fast}$} & None & None & None & True & True & True & False & False & False \\ 
\textbf{$l_{\rm w}$} & 0.03 & 0.03 & 0.03 & 0.005 & 0.005 & 0.005 & 0.0025 & 0.01 & 0.01 \\ 
\textbf{$\lambda_{ksp}$} & 5 & 5 & 5 & 10 & 10 & 10 & 5 & 5 & 5 \\ 
\textbf{$\lambda_{L_1}$} & 1 & 1 & 1 & 0.1 & 0.1 & 0.1 & 1 & 1 & 1 \\ 
\textbf{$\lambda_H$} & 1 & 1 & 1 & 2 & 2 & 2 & 1 & 1 & 1 \\ 
\textbf{Weight init} & 0.1 & 0.1 & 0.1 & 0.1 & 0.1 & 0.1 & 0.1 & 0.1 & 0.1 \\ 
\textbf{$\sigma$} & 0.01 & 0.01 & 0.01 & 0.01 & 0.01 & 0.01 & 0.01 & 0.01 & 0.01 \\ 
\bottomrule
\end{tabular}
\caption{Hyperparameters for Split-MNIST, Split-CIFAR10, and Tiny-ImageNet datasets. "None" indicates that the dual memory architecture is not involved; "\protect${\mathcal{D}}^{\rm slow}_b\protect$" and "\protect${\mathcal{D}}^{\rm fast}_b\protect$" represent the number of samples selected from the slow memory and fast memory, respectively, during memory replay. For example, if "\protect${\mathcal{D}}^{\rm slow}_b\protect$" = 0.5, it means that "0.5 * batch\_size" samples are selected from the slow memory for memory replay. The batch\_size parameter of ITDMS is set to be the same as that of DER++ in the same scenario; "True" means that the fast memory needs to be reset after completing the current task's training and before moving on to the next task, while "False" means that no reset operation is performed; "Weight init" indicates the initial value of the sample weight, which is fixed at 0.1 in all training scenarios.}
\end{table}

\begin{table}[h]
\centering
    \fontsize{9}{11}\selectfont
\begin{tabular}{c ccc ccc} 
\toprule
\textbf{Data set} & \multicolumn{3}{c}{\textbf{Perm-MNIST}} & \multicolumn{3}{c}{\textbf{Rot-MNIST}} \\ 
\cmidrule(r){2-4} \cmidrule(r){5-7} 
\textbf{Memory size} & 200 & 500 & 1000 & 200 & 500 & 1000 \\ 
\midrule
\textbf{$\lambda_{H_2}$} & -1 & -1 & -1 & -1 & -1 & -1 \\ 
\textbf{$\lambda_{CS}$} & 1 & 1 & 1 & 1 & 1 & 100 \\ 
\textbf{$|{\mathcal{M}}^{\rm slow}|$} & 200 & 300 & 500 & 200 & 300 & 500 \\ 
\textbf{$|{\mathcal{M}}^{\rm fast}|$} & None & 200 & 500 & 200 & 200 & 500 \\ 
\textbf{${\mathcal{D}}^{\rm slow}_b$} & 1.0 & 0.6 & 0.5 & 1.0 & 0.6 & 0.5 \\ 
\textbf{${\mathcal{D}}^{\rm fast}_b$} & 1.0 & 0.4 & 0.5 & 1.0 & 0.4 & 0.5 \\ 
\textbf{Reset ${\mathcal{M}}^{\rm fast}$} & None & False & False & True & False & False \\ 
\textbf{$l_{\rm w}$} & 0.005 & 0.005 & 0.005 & 0.005 & 0.005 & 0.005 \\ 
\textbf{$\lambda_{ksp}$} & 10 & 10 & 10 & 10 & 10 & 10 \\ 
\textbf{$\lambda_{L_1}$} & 1 & 1 & 1 & 1 & 1 & 1 \\ 
\textbf{$\lambda_H$} & 1 & 1 & 1 & 1 & 1 & 1 \\ 
\textbf{Weight init} & 0.1 & 0.1 & 0.1 & 0.1 & 0.1 & 0.1 \\ 
\textbf{$\sigma$} & 0.01 & 0.01 & 0.01 & 0.01 & 0.01 & 0.01 \\ 
\bottomrule
\end{tabular}
\caption{Hyperparameters for Perm-MNIST and Rot-MNIST datasets.}
\end{table}

\begin{figure}[t]
    \centering
    \includegraphics[trim=1.5cm 1cm 2.1cm 1cm, clip, width=1.0\textwidth, page=8]{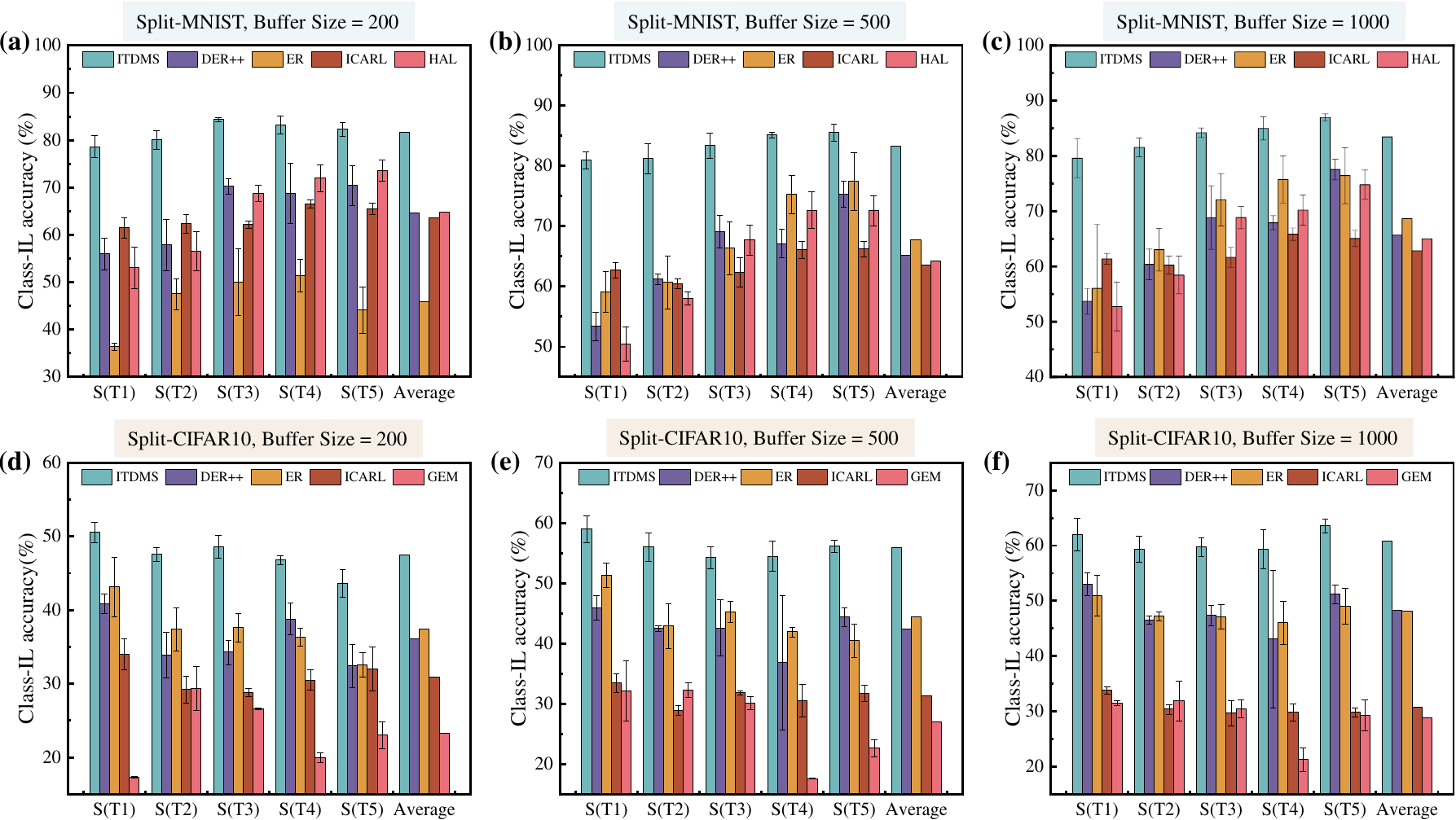}
    \caption{Performance comparison between the proposed ITDMS framework and baselines in imbalanced data stream scenarios. All results are averaged over 10 runs. (a), (b), and (c)~: The results of various models on the Split-MNIST dataset using the memory sizes of 200, 500, and 1000, respectively. (d), (e), and (f)~: The results of various models on the Split-CIFAR10 dataset using the memory sizes of 200, 500, and 1000, respectively.}
    \label{fig:imbalacnedResults}
\end{figure}

\subsection{Experiment Results on the Imbalanced Data Stream}

In Section~\ref{sec:balacnedResults}, we evaluate the effectiveness of our proposed ITDMS framework under balanced data stream conditions by comparing its continual learning performance against various baseline methods across four different datasets. However, in real-world continual learning scenarios, the data stream usually contains a different number of data samples per class. Learning such an imbalanced data stream remains a challenge since one category would contain more data samples than other categories in each task. As a result, the model trained on such a data stream tends to perform well in the majority class and leads to performance degeneration in the other classes.

To relieve this issue, ensuring a relatively balanced distribution of historical data in memory proves to be an effective strategy. The proposed ITDMS model can optimize the sample selection weights to obtain a nearly equal number of samples from each category. In addition, the proposed memory optimization mechanism, defined by Eq.~\eqref{weightOptimization2} can also ensure that the selected data samples contain diverse information for each category. Moreover, the proposed BBS approach can further ensure the category balance in the slow memory buffer by using Eq.~\eqref{sampleRemove} that automatically removes an appropriate number of redundant memorized samples per category. After the sample removal process, the number of remaining samples for each category is equal. Consequently, our proposed memory system can work better and achieve super performance in imbalanced data stream scenarios when compared with other baselines that can not address the imbalanced data samples.

We evaluate the performance of the proposed ITDMS framework and compare it with other baselines in imbalanced data stream scenarios. Specifically, we consider two datasets~: Split-MNIST and Split-CIFAR10. The first task in the imbalanced data stream has 4,000 data samples and the other task has 400 samples, which maintains an imbalance ratio of 10:1. We call this learning setup as S(Ti). We consider training various models with three memory configurations, including 200, 500 and 1,000, respectively.

The classification accuracy of various models on the imbalanced data stream is presented in Fig.~\ref{fig:imbalacnedResults}. From the results, we can find that the proposed ITDMS framework consistently exhibits superior performance and can maintain stable performance across all tasks and continual learning settings compared to other baseline methods, particularly in scenarios with imbalanced data streams. This advantage is primarily attributed to the proposed memory optimization approach. In contrast, the other memory-based methods such as DER++ and ER employ a random sample process to update the memory buffer, which can perform well in the balanced data stream but suffers from significant performance degeneration in the imbalanced data stream. The iCaRL is also a popular memory-based approach, which typically selects samples closest to the class centre to maintain intra-class representativeness and inter-class separation. However, the iCaRL fails to allocate representative samples fairly for each class and cannot effectively update the class centre for minority classes, resulting in a highly skewed and unrepresentative distribution in memory. In imbalanced data streams, minority-class data may be overlooked or deemed less important by HAL, making it difficult for such data to be selected as anchor samples. Consequently, this leads to insufficient representation of minority class data in memory, causing HAL to forget these minority classes more easily.

Regarding GEM, while its gradient constraint mechanism can help mitigate the negative impact of new task training on the performance of previous tasks—even under imbalanced data—this mechanism only reduces catastrophic forgetting by directly limiting the damage caused by new tasks to old ones. Although GEM can retain some memory of minority class samples to a certain extent, its mechanism is not inherently designed to handle imbalanced data scenarios. It lacks strategies to adjust memory allocation to support minority classes and does not adapt gradient updates based on class imbalance, thus failing to proactively address the problem of class imbalance.

\subsection{Analysis Results}
\begin{figure}[t]
    \centering
    \includegraphics[trim=0cm 2.5cm 0cm 1.8cm, clip, width=1.0\textwidth, page=4]{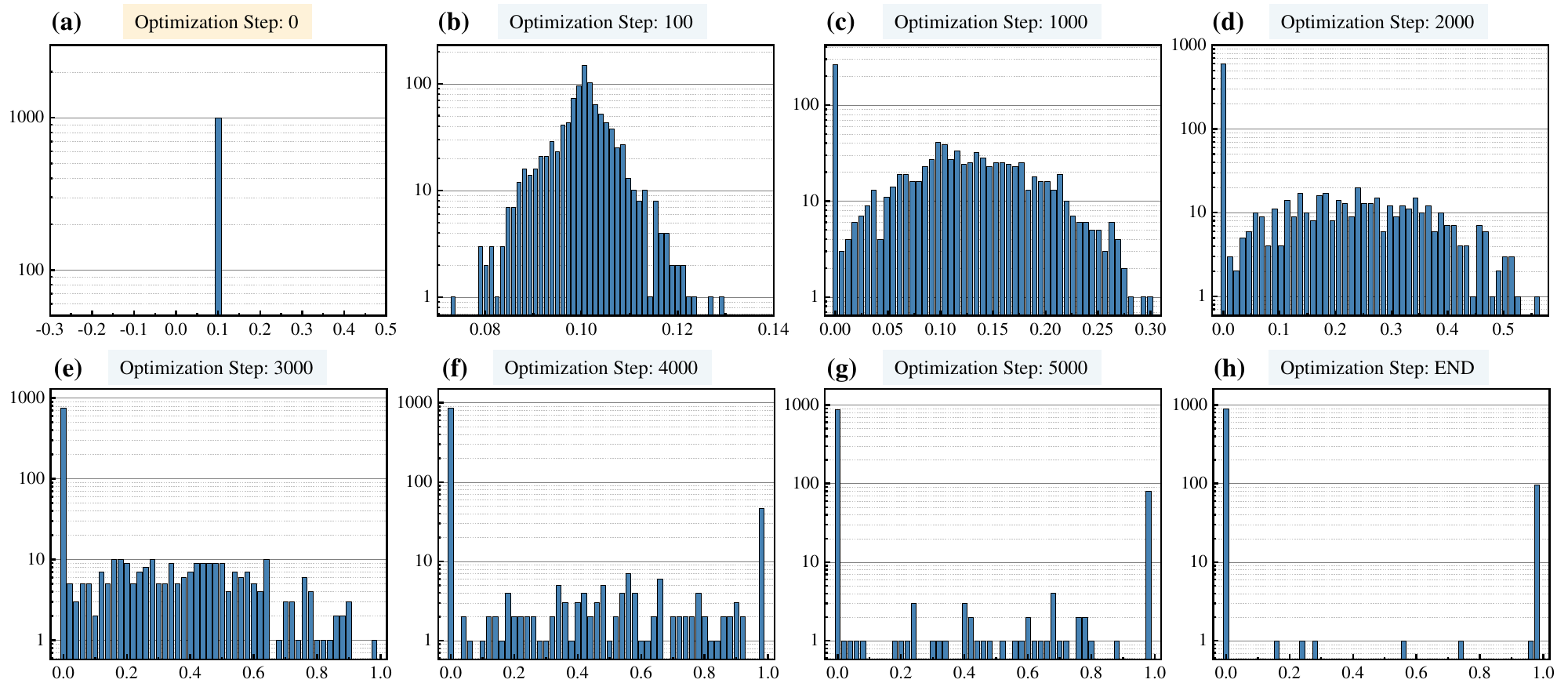}
    \caption{The distribution changes of the sample selection weights during the optimization process. The horizontal and vertical coordinates represent the value distribution of sample weights and their corresponding quantity proportions, respectively: (a) The results when all weights are initialized to 0.1, indicating that each sample is assigned by the same sample selection weight; (b-g) The distribution changes of the sample selection weights, which goes through a global search and weight sparsification phase, indicating that the model is evaluating the importance of selected samples, eventually converging, as shown in (h), with the majority of data points close to 0 and a few data points close to 1.}
    \label{fig:4}
\end{figure}

In the first, we construct an empirical experiment to show the property of the proposed memory system. We consider creating a Gaussian mixture distribution dataset, which consists of 1,000 data points categorized into four classes, each following a mixture Gaussian distribution with relatively fixed spatial locations. This dataset can simulate the distributional differences between the classes for a general dataset. We report the results in Fig.~\ref{fig:4}. The dispersion of the original data points represents the overall diversity of the dataset, while the density between different groups indicates latent information characteristics. Our goal is for the model to identify high-value samples that are both diverse and representative of the overall dataset distribution by optimizing the sample weights.

\begin{figure}[h!]
    \centering
    \includegraphics[trim=0.1cm 5.7cm 0.1cm 4.6cm, clip, width=1.0\textwidth, page=3]{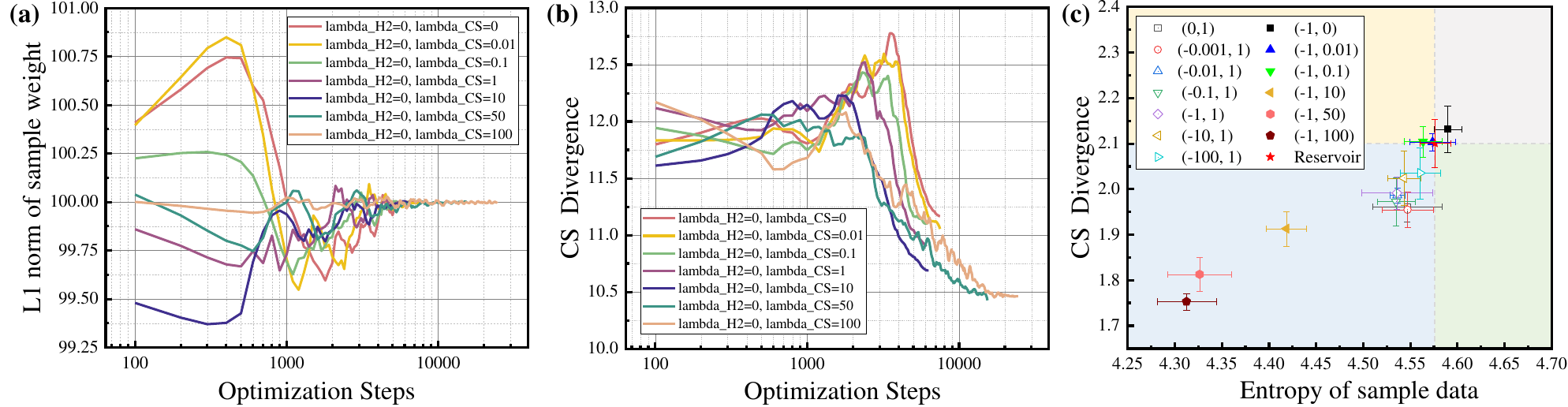}
    \caption{Changes of variables under different $\lambda_{H2}$ and $\lambda_{CS}$ settings throughout the optimization steps: (a) The L1 norm of sample selection weights undergoes an initial oscillation process and stabilizes after the 1000-th optimization step; correspondingly, (b) The change on the CS divergence between the sampled points and the original data points; (c) The effectiveness of the samples selected under different parameter settings relative to the random selection method, where the x-axis represents the entropy (diversity) of the selected samples, and the y-axis represents the CS divergence (representativeness) of the selected samples relative to the original dataset. The legend coordinates represent the values of $\lambda_{H2}$ and $\lambda_{CS}$ (e.g., (-1,1) represents $\lambda_{H2}$ = -1, $\lambda_{CS}$ = 1).}
    \label{fig:5}
\end{figure}

In the simulation results, Fig.~\ref{fig:5} effectively illustrates the changes in sample weight distribution during the optimization process. Assuming that we want the model to select 100 high-value sample points, the sample weights are initially set to the same constant at the beginning of each optimization process, ensuring equal importance for all samples and preventing the algorithm from converging to suboptimal solutions due to random initial weight assignments. As the optimization progresses, the sample weights are gradually clustered into several parts then back to a concentrated cluster. This indicates that the model optimizes the weights to identify an optimal subset of samples, in which the weights for most of the unimportant samples are near zero while the weights for the remaining samples approximate 1.

To further explain the internal optimization phase of the proposed memory system, we adjust the hyperparameters of second-order Rényi entropy ($\lambda_{H2}$) and CS divergence ($\lambda_{CS}$) in the loss function. For normalization and control variable considerations, we fixed $\lambda_{H2}$ at -1 or $\lambda_{CS}$ at 1. Fig.~\ref{fig:5}(a) shows that the L1 norm of sample weights fluctuates during early training steps and gradually converges to a fixed value, equivalent to the sample number, after 1,000 optimization steps. Interestingly, the corresponding CS divergence between the selected subset and original datasets increases in the first phase and then gradually decreases in the subsequent optimization steps, even when the sum of the sample weights has stabilized. Additionally, we provide the distribution change for the sample selection weight in Fig.~\ref{fig:4}. These results show that even after achieving the k-sparse constraint, the model persistently attempts to adjust the internal distribution of sample weights to minimize the difference between the selected subset and original dataset distributions, indicating an effort to select a subset of samples that better represents the entire sample space, rather than merely optimizing within the constraints of the weight parameters.

\begin{figure}[h!]
    \centering
    \includegraphics[trim=7.5cm 1.8cm 9cm 1.7cm, clip, width=1.0\textwidth, page=5]{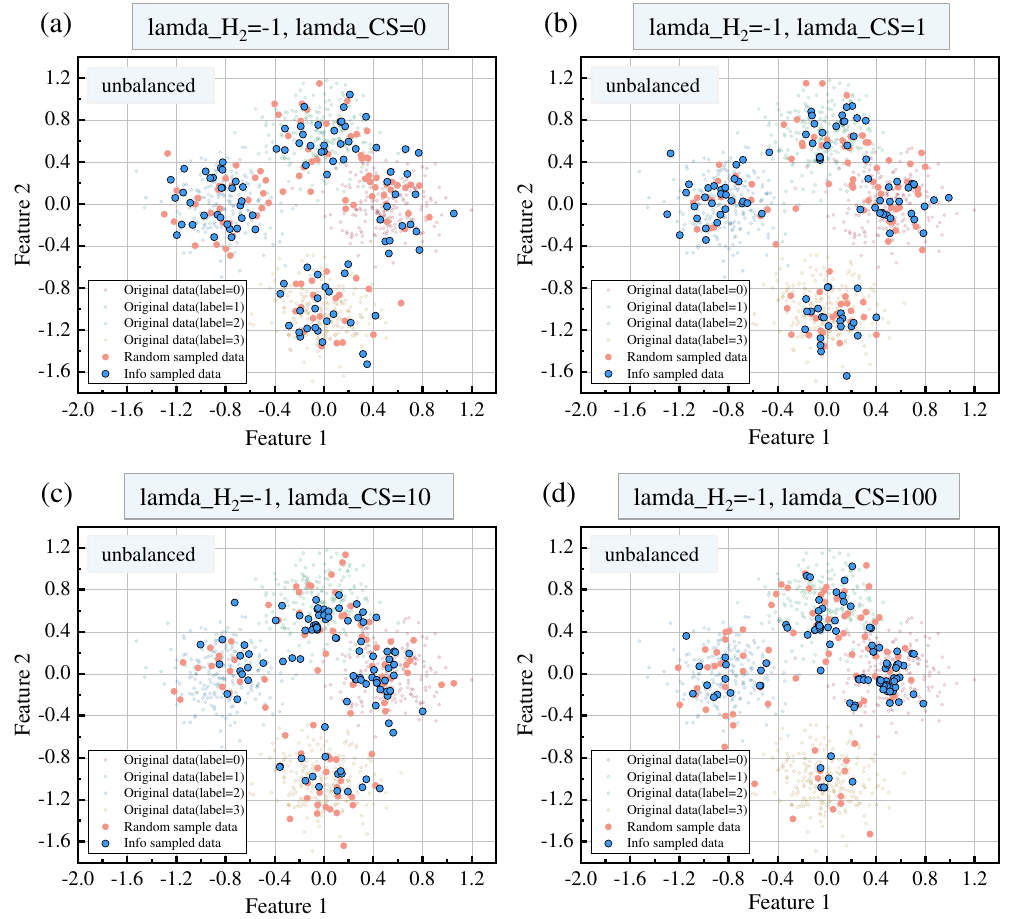}
    \caption{Comparison of sampling effects between unbalanced sampling mode against the random sampling method (reservoir strategy), The terms Feature 1 and 2 refer to the features of the two-dimensional simulated Gaussian distribution dataset: In unbalanced sampling modes, when $\lambda_{CS}$ equals 0, the sampling effect is consistent with the random method. As $\lambda_{CS}$ increases, the samples selected by ITDMS tend to reflect the distribution characteristics and data structure of the samples more intensively. However, when there is multi-label data in the task, the unbalanced mode tends to select different amounts of samples for each label. The results for the balanced setting are shown in Figure.~\ref{fig:7} }
    \label{fig:6}
\end{figure}

\begin{figure}[h!]
    \centering
    \includegraphics[trim=7.6cm 1.9cm 9.0cm 1.7cm, clip, width=1.0\textwidth, page=6]{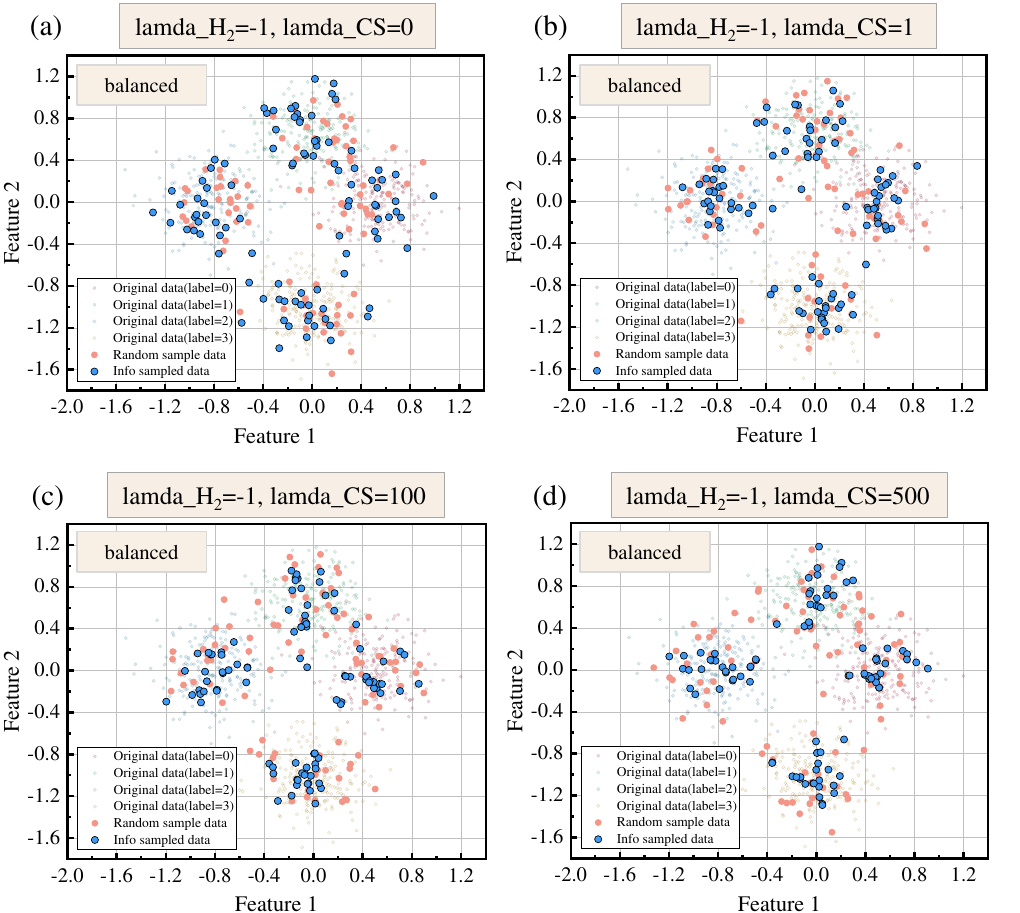}
    \caption{Comparison of sampling effects between the balanced sampling mode against the random sampling method (reservoir strategy).}
    \label{fig:7}
\end{figure}

From the results of Fig.~\ref{fig:5}(c), we can observe that the trade-offs between diversity and representativeness of the selected subset at the final optimization stage, are influenced by the different $\lambda_{H2}$ and $\lambda_{CS}$ configurations. We use the entropy of randomly selected samples and their CS divergence from the original dataset as reference points, dividing the entire range into four quadrants. Ideally, we hope the model-selected points fall in the fourth quadrant. However, due to the directional nature of the sampling process, the model is less likely to select a sample set with higher diversity compared to random sampling—unless $\lambda_{CS} = 0$, which implies sacrificing representativeness to achieve higher diversity. Thus, although our sampling results are mostly distributed in the first and third quadrants—where the latter sacrifices some diversity to capture more of the dataset's structural characteristics, and the former does the opposite, which still reflects a reasonable and effective model performance. The absence of results in the second quadrant indicates that simultaneously sacrificing diversity and failing to capture latent structural features is the worst outcome, which should be avoided at all costs. Furthermore, we observed that the impact of $\lambda_{H2}$ is far less significant than that of $\lambda_{CS}$, leading us to primarily adjust $\lambda_{CS}$ while keeping $\lambda_{H2}$ constant at -1 during the actual optimization process.

The proposed approach has two sampling strategies: global sampling and balanced sampling based on the class labels. The former tends to capture the global characteristics of the data but may result in unbalanced sampling. When training on the Split-MNIST and Split-CIFAR10 datasets, each task contains only two label categories, making this imbalance effect less noticeable. When the label distribution in the training set is complex and the buffer capacity is extremely limited, the balanced sampling can achieve good results. For example, when we consider learning a model with a memory buffer of size of 200 on the Sequential-Tiny ImageNet, using the balanced sampling can ensure that the memory buffer stores at least one sample for every category. However, other baselines that employ a random selection strategy for updating the memory buffer can not store all category samples for the Sequential-Tiny ImageNet using a small-size memory buffer, leading to performance degeneration.

The above discussion explains why ITDMS outperforms other models by over $5\%$ on the Sequential-Tiny ImageNet when the buffer size is 200, as shown in Table.~\ref{tab:averAccuracy}. Fig.~\ref{fig:6} clearly shows the difference in sample selection balance between the two sampling strategies. We observe that as $\lambda_{CS}$ increases, sampled points become more concentrated around the data's structural characteristics and density. The balanced sampling strategy can avoid the label imbalance that may occur in the unbalanced continual learning setting, and this strategy was consistently applied across various continual learning settings.

\begin{figure}[h!]
    \centering    \includegraphics[trim=6cm 0cm 6.2cm 0cm, clip, width=1.0\textwidth, page=9]{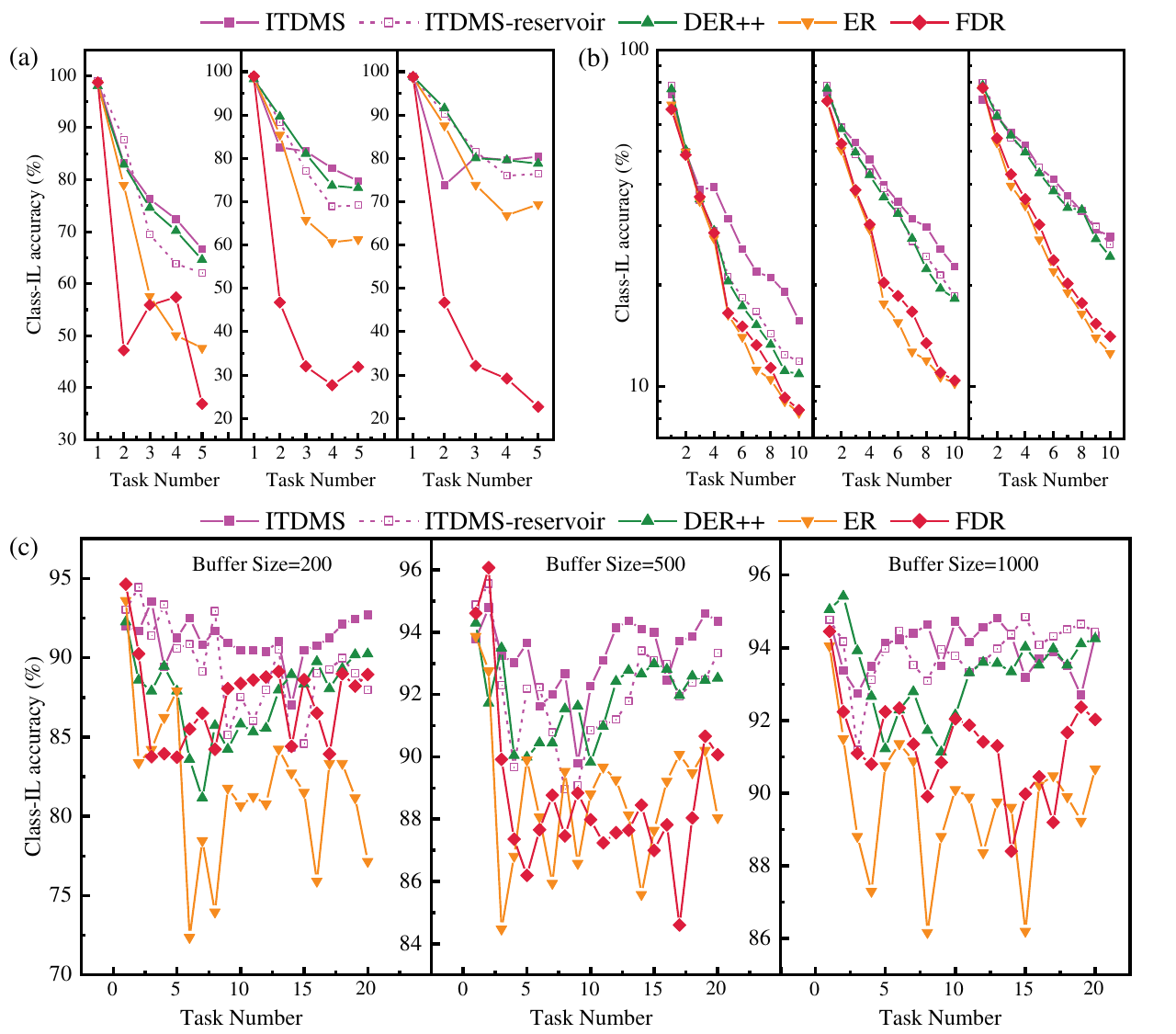}
    \caption{The results of the ablation study. (a), (b), and (c) ~: The classification accuracy curves of the proposed ITDMS framework and the control group on the Class-IL task sequence for the Split-CIFAR10, Sequential-Tiny ImageNet, and R-MNIST datasets. For each dataset, we consider three memory capacity configurations (200, 500, and 1000). The "ITDMS-reservoir" represents the baseline that uses the reservoir sample to update the fast and slow memory buffer. 
    }
    \label{fig:differentSampleing}
\end{figure}

\subsection{Ablation Experiment}
To achieve a more nuanced comprehension of the model's internal workings while improving its interpretability and transparency, we consider constructing an ablation study aimed at investigating and comparing the impact of the proposed memory optimization approach on the model's performance. First, we consider creating a baseline model that employs the ITDMS memory system but replaces the proposed memory optimization approach using the reservoir sampling, namely ITDMS-reservoir. In addition, we also do not use the proposed BBS strategy to maintain balanced samples for the ITDMS-reservoir. As a result, the ITDMS-reservoir employs two memory buffers, which are based on reservoir sampling. Furthermore, we employ the same hyperparameter configuration for both ITDMS and ITDMS-reservoir.

We train all models with three different memory configurations (200, 500, and 1000) on three datasets, including Split-CIFAR10, Sequential-Tiny ImageNet and R-MNIST. We report the results in Figure.~\ref{fig:differentSampleing}, which shows that the baseline model (ITDMS-reservoir) generally underperforms compared to the proposed ITDMS framework in most scenarios. These results demonstrate that the proposed memory optimization approach can improve the model's performance, 
especially in the imbalanced data stream where the number of samples per task differs.

However, we can also find that the proposed ITDMS framework can only achieve more significant performance improvement than the ITDMS-reservoir when using a small memory size such as 200. In contrast, when using a large memory size such as 1000, the baseline model (ITDMS-reservoir) outperforms the DER++ and achieves similar performance on the Sequential-Tiny ImageNet and R-MNIST compared to the proposed ITDMS framework. These results show that using the dual memory system with the reservoir sampling can still achieve better results than the DER++ which is based on a single memory system. 

The proposed ITDMS framework exhibits a consistent superiority over the baseline model (ITDMS-reservoir) across various memory configurations, with the disparity in performance becoming more significant at small memory capacities compared to large ones. Three key insights can be drawn from the results: First, the dual memory architecture reveals a greater capability to reduce forgetting than a single memory system (such as DER++ and ER) utilizing the same sample selection methodology. Second, the proposed memory optimization technique further enhances the dual memory system's ability to counteract network forgetting in continual learning, as evidenced by the results illustrated in Fig.~\ref{fig:differentSampleing}. Lastly, the ITDMS memory system significantly surpasses the baseline model when using limited memory conditions, highlighting its practical utility in scenarios where memory resources are critically constrained.

\section{Conclusion and Future Works}
The results derived from the CLS theory framework suggest that information is processed through both rapid and gradual learning mechanisms. Building on this foundation, we propose the implementation of these mechanisms via the Information-Theoretic Dual Memory System (ITDMS), which comprises both a slow and a fast memory buffer. The fast memory buffer utilizes a reservoir sampling technique to dynamically replace outdated memorized samples with new data. Additionally, this paper introduces an information-theoretic memory optimization strategy that assesses the quality of each sample according to an information cost function, offering a systematic approach for selecting and retaining the most critical data in the slow memory buffer. Moreover, we present a novel balanced sample selection method that allows the memory system to flexibly allocate memory capacity for the storage of new samples. Empirical results from a series of experiments illustrate that the proposed approach, when integrated into existing continual learning models, can significantly enhance their performance.

The fundamental constraint of the proposed method lies in its inability to manage an unbounded array of tasks due to the limitations inherent in both the model and memory capacity. To mitigate this issue, one viable approach is to introduce an innovative dynamic memory expansion strategy that progressively enhances the slow memory buffer's capacity in alignment with the data stream's complexity. Furthermore, an alternative solution involves the development of a novel dynamic model expansion strategy that methodically constructs a new sub-model within a mixture framework. Specifically, this dynamic model expansion strategy utilizes the novelty of a task as a triggering signal for expansion, thereby promoting an efficient network architecture.

One significant drawback of the proposed methodology is that the information-theoretic memory optimization strategy necessitates substantial computational resources, as it requires numerous optimization iterations following each transition between tasks. To address this challenge, we aim to devise an innovative acceleration technique that minimizes the number of iterations involved in the memory optimization process, ensuring the retention of critical and informative data samples. This approach will be systematically developed and explored in my forthcoming research.
\bibliographystyle{elsarticle-num}
\bibliography{VAEGAN}

\end{document}